\documentclass[journal]{IEEEtran}
\usepackage{bm}
\usepackage{amsfonts}
\usepackage{amsmath}

\usepackage{graphicx}  
\usepackage{multirow}
\usepackage{booktabs}
\begin{document}

\title{Adversarial Feature Learning of Online Monitoring Data for Operational Risk Assessment in Distribution Networks}

\author{ Xin~Shi$^1$,
         Robert~Qiu$^{1,2}$,~\IEEEmembership{Fellow,~IEEE},
         Tiebin~Mi$^1$,
         Xing~He$^1$,
         and~Yongli~Zhu$^3$

\thanks{This work was partly supported by National Key R \& D Program of No. 2018YFF0214705, NSF of China No. 61571296, (US) NSF Grant No. CNS-1619250 and NSF of China No. 51677072.

$^1$ Department of Electrical Engineering,Center for Big Data and Artificial Intelligence, State Energy Smart Grid Research and Development Center, Shanghai Jiaotong University, Shanghai 200240, China. (e-mail: dugushixin@sjtu.edu.cn; rcqiu@sjtu.edu.cn; mitiebin@foxmail.com; hexing\_hx@126.com)

$^2$ Department of Electrical and Computer Engineering,Tennessee Technological University, Cookeville, TN 38505, USA. (e-mail:rqiu@tntech.edu)

$^3$ State Key Laboratory of Alternate Electrical Power Systems with Renewable Energy Sources, North China Electric Power University, Baoding 071003, China. (e-mail: yonglipw@163.com)
}}


\maketitle

\begin{abstract}
With the deployment of online monitoring systems in distribution networks, massive amounts of data collected through them contains rich information on the operating states of the networks. By leveraging the data, an unsupervised approach based on bidirectional generative adversarial networks (BiGANs) is proposed for operational risk assessment in distribution networks in this paper. The approach includes two stages: (1) adversarial feature learning. The most representative features are extracted from the online monitoring data and a statistical index $\mathcal{N}_{\phi}$ is calculated for the features, during which we make no assumptions or simplifications on the real data. (2) operational risk assessment. The confidence level $1-\alpha$ for the population mean of the standardized $\mathcal{N}_{\phi}$ is combined with the operational risk levels which are divided into emergency, high risk, preventive and normal, and the p value for each data point is calculated and compared with $\frac{\alpha}{2}$ to determine the risk levels. The proposed approach is capable of discovering the latent structure of the real data and providing more accurate assessment result. The synthetic data is employed to illustrate the selection of parameters involved in the proposed approach. Case studies on the real-world online monitoring data validate the effectiveness and advantages of the proposed approach in risk assessment.
\end{abstract}

\begin{IEEEkeywords}
operational risk assessment, distribution networks, online monitoring data, bidirectional generative adversarial networks (BiGANs), unsupervised approach
\end{IEEEkeywords}

%
\IEEEpeerreviewmaketitle
\section{Introduction}
\label{section: Introduction}
\IEEEPARstart{T}{he} operational risk assessment is a fundamental task in distribution networks, which can help realize situation awareness of the network and offer support on the safety analysis and control decision. One main factor that influences the operational risks are faults or fluctuations caused by anomalies in the distribution network. These anomalies may present intermittent, asymmetric, and sporadic spikes, which are random in magnitude and could involve sporadic bursts as well, and exhibit complex, nonlinear, and dynamic characteristics \cite{jaafari2007underground}. Additionally, with numerous branch lines and changeable network topology, it is questionable for the traditional model-based approaches to fully and accurately detect the anomalies in the distribution network, because they are usually based on certain assumptions or simplifications.

In recent years, there have been significant deployments of online monitoring systems in distribution networks and a large amount of data is collected through them. The massive data contains rich information on the operating state of the distribution network. In order to leverage the data, many advanced analytics are developed. For example, in \cite{xie2014dimensionality}, a PCA-based approach is proposed to reduce the dimensionality of phasor-measurement-unit (PMU) data, whose result is utilized to detect early events in the network. In \cite{adewole2012fault}, based on wavelet energy spectrum entropy decomposition of disturbance waveforms, characteristic features are extracted to detect and classify faults in a distribution network. In \cite{wu2017online}, by using multi-time-instant synchrophasor data, a density-based outlier detection approach is used to detect low-quality synchrophasor measurements. In \cite{pignati2017fault}, by computing parallel synchrophasor-based state estimators, a real-time fault detection and faulted line identification methodology is proposed. In \cite{ghaderi2015high}, a method using time-frequency analysis is proposed for feature extraction, and a classifier is trained by those extracted features. In \cite{Chu2016Massive}, based on the multiple high dimensional covariance matrix test theory, a statistically-based anomaly detection algorithm is proposed for streaming PMU data.

Reviewing the current efforts on anomaly detection by using online monitoring data, two main weaknesses exist: 1) they rely on a prior parametric model of the monitoring data, which are usually based on certain assumptions and simplifications. 2) they often use simple features calculated through the time-series monitoring data, such as the mean, variance, spectrum, high moments, etc. For the methods based on pre-designed parametric models, they are sensitive to parameter values and it is not easy to find the optional parameters to capture the essential features for each data segment. Therefore, false alarm can easily happen. While for the statistically-based methods, simple statistical features often are not well generalized in most cases and they are susceptible to random fluctuations, which makes it impossible to detect the latent anomalies.

Generative adversarial nets(GANs) are first proposed by Goodfellow in 2014 \cite{Goodfellow2014Generative}, overcoming the difficulties of approximating many intractable probabilistic computations and leveraging the benefits of piecewise linear units in deep generative models. It trains a generator to automatically capture the distribution of the sample data from simple latent distributions, and a discriminator to distinguish between real and generated samples. Compared with the probability distributions calculated by traditional techniques, the automatically captured ones can better depict the rich structure information of the arbitrarily complex data. However, the existing GANs have no function of projecting the generated data back into the latent space, which makes it impossible to use those latent feature representations for auxiliary problems. In view of this occasion, Donahue propose an improved framework of GANs in 2016, bidirectional generative adversarial networks (BiGANs)\cite{DBLP:journals/corr/DonahueKD16}. BiGANs have solved the problem of inaccessible feature representations by adding an encoder in the original framework of GANs. It is a robust and highly generative feature learning approach for arbitrarily complex data, making no assumptions or simplifications on the data.

In this paper, we propose a new unsupervised approach for the operational risk assessment in distribution networks. It can automatically learn the most representative features from the input data in an adversarial way by using BiGANs. Based on the extracted features, a statistical index $\mathcal{N}_{\phi}$ is defined and calculated to indicate the data behavior. Furthermore, to quantify the operational risks of feeder lines in the distribution network, the risk levels are classified into emergency, high risk, preventive and normal, and they are combined with the confidence level $1-\alpha$ for the population mean of the standard $\mathcal{N}_{\phi}$. By comparing the p value for each data point of the standard $\mathcal{N}_{\phi}$ with the intervals of $\frac{\alpha}{2}$, the operational risks of the feeder lines can be judged intuitively. The main advantages of the proposed approach are summarized as follows:
1) It is a purely data-driven approach without requiring too much prior knowledge on the complex topology of the distribution network, which eliminates the potential detection errors caused by inaccurate network information.
2) It is an unsupervised learning approach requiring no anomaly labels or records, which solves the label lack or inaccuracy problem in the distribution network.
3) It automatically learns features from the online monitoring data in the distribution network, which makes it possible for detecting the latent anomalies. Because the learned features are more powerful in representing the real data than artificially designed ones.
4) It is suitable for both online and offline analysis.

The rest of this paper is organized as follows. Section \ref{section: theory} describes the proposed BiGANs-based anomaly detection algorithm, i.e., data preparing and normalization, adversarial feature learning, and anomaly detection. In section \ref{section: application}, spatio-temporal matrices are formulated by using the online monitoring data in distribution networks and specific steps of operational risk assessment are given. In section \ref{section: case}, the synthetic data from IEEE 118-bus system are used to illustrate the selection of parameters involved in the approach, and the real-world online monitoring data in a distribution network are used to validate the effectiveness and advantages of the proposed approach. Conclusions are presented in section \ref{section: conclusion}.
\section{BiGANs-Based Anomaly Detection}
\label{section: theory}
In this section, the BiGANs-based anomaly detection algorithm is introduced. First, the multi-dimensional time series data are partitioned into a series of segments in chronological order. The main idea is that BiGANs are utilized to automatically learn the most representative features from the data segments without making any prior assumptions or simplifications. Based on the extracted features for each data segment, a statistical index is calculated to indicate the data behaviour. The designed algorithm offers an end-to-end solution for anomaly detection and specific steps are characterized as below.
\subsection{Data Preparing and Normalization}
\label{subsection: data preparing and normalization}
Assume there are $P$-dimensional measurements (such as voltage measurements from $P$ sensors installed on one feeder line) $(d_1,d_2,...,d_P)\in \mathbb{R}^P$. At the sampling time $t_j$, the $P-$dimensional measurements can be formulated as a column vector ${\bf d}(t_j)=(d_1,d_2,...,d_P)^H$. For a series of time $T$, by arranging these vectors in chronological order, a spatio-temporal data matrix ${\bf D}\in \mathbb{R}^{P\times T}$ is obtained.

\begin{figure}[!t]
\centering
\includegraphics[width=2.5in]{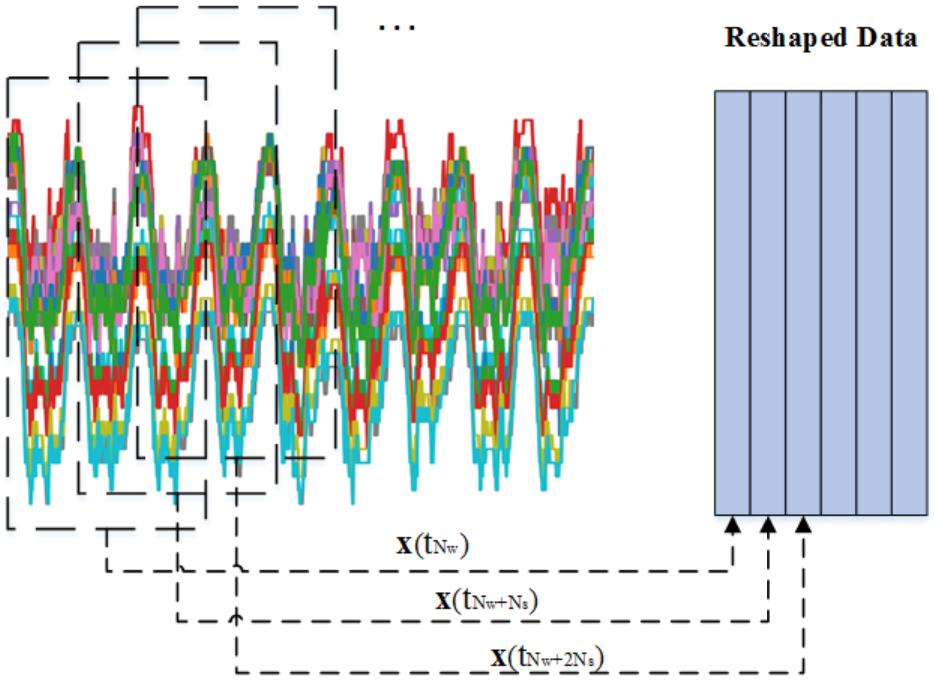}
\caption{Pipeline for the vectorization of data matrix.}
\label{fig:data_vectorization}
\end{figure}

With a $P\times {N_w}$ (${N_w}\le T$) window moving on ${\bf D}$ at the step size $N_s$, a series of data segments are generated. For example, at the sampling time $t_j$, the generated data segment is
\begin{equation}
\label{Eq:data_form}
\begin{aligned}
  {{\bf D}}(t_j) = \left( {{{\bf d}}(t_{j - {N_w} + 1}),{{\bf d}}(t_{j - {N_w} + 2}), \cdots ,{{\bf d}}(t_j)} \right)
\end{aligned},
\end{equation}
where ${{\bf d}}(t_k)={({d_1,d_2,\cdots,d_P})}^H$ ($t_{j-{N_w}+1}\le t_k \le t_j$) is the sampling data at time $t_k$. For the data segment at the sampling time $t_j$, we reshape it into a column vector denoted as ${\bf x}_{t_j}\in \mathbb{R}^{P{N_w}\times 1}$. Thus, the spatio-temporal data matrix is reformulated as ${{\bf D}}=[{\bf x}({t_{N_w}});{\bf x}({t_{{N_w}+N_s}});{\bf x}({t_{{N_w}+2\times N_s}});\cdots]$, which is shown in Figure \ref{fig:data_vectorization}.
To reduce the calculation error and improve the convergence speed of training BiGANs in the subsequent feature learning process, we normalize $\bf x$ into $\tilde{\bf x}$ by
\begin{equation}
\label{Eq:min_max_normalize}
\begin{aligned}
  {\tilde x_{i}} = \frac{{{x_{i}} - \min ({{\bf{x}}})}}{{\max ({{\bf{x}}}) - \min ({{\bf{x}}})}}
\end{aligned},
\end{equation}
where ${\tilde x_{i}}(i = 1,2,\cdots,PN_w)$ is the normalized value in the range $[0,1]$, $\min({{\bf{x}}})$ and $\max({{\bf{x}}})$ respectively denote the minimum and maximum value of $\bf x$.
\subsection{Adversarial Feature Learning}
\label{subsection: adversarial_feature_learning}
\begin{figure}[!t]
\centerline{
\includegraphics[width=2.5in]{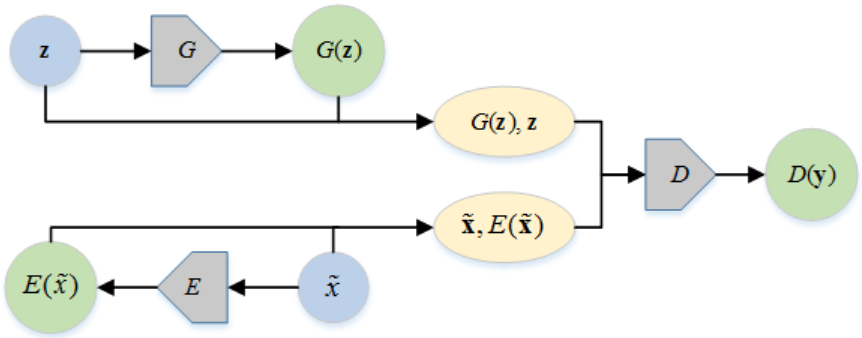}
}
\caption{The framework of BiGANs.}
\label{fig:BiGANs}
\end{figure}

BiGANs, train a generative model $G$ to capture the distribution of the sample data, a discriminative model $D$ to distinguish the real samples from the generated ones as accurately as possible, and an encoding model $E$ to project sample data back into the latent space. Feature representations learned by the encoding model can depict the rich structure information of the sample data well, which can be used for other auxiliary problems. The framework of BiGANs is shown in Figure \ref{fig:BiGANs}.

The generative model $G$, also called generator, is composed of multi-layer networks. For convenience, we assume it's a three-layer network, i.e., the input layer, one hidden layer and the output layer. Data transition from the input layer to the output layer can be denoted as in Equation (\ref{Eq:g_input_hidden}) and (\ref{Eq:g_hidden_output}).
\begin{equation}
\label{Eq:g_input_hidden}
\begin{aligned}
  {\bf h}_{g} = {S_h}({\bm W}_{h}{{\bf z}}+{{\bm b}_h})
\end{aligned},
\end{equation}
\begin{equation}
\label{Eq:g_hidden_output}
\begin{aligned}
  G({\bf z}) = {S_g}({\bm W}_{g}{{\bf h}_g}+{{\bm b}_g})
\end{aligned},
\end{equation}
where $\bf z$ is the input data sampled from a simple latent distribution (gaussian, uniform, exponential, etc), ${\bf h}_{g}$ is the output of the hidden layer in $G$, ${\bm W}_h$ is the weight matrix between the input layer and the hidden layer and ${\bm W}_g$ is the weight matrix between the hidden layer and the output layer. To avoid gradient vanishing in training multi-layer $G$, as proposed in \cite{glorot2010understanding}, ${\bm W}_h$ and ${\bm W}_g$ can be initialized by using an uniform distribution
\begin{equation}
\label{Eq:weight_initialization}
\begin{aligned}
  U[-\frac{\sqrt 6}{\sqrt {{M_i}+{M_{i+1}}}}, \frac{\sqrt 6}{\sqrt {{M_i}+{M_{i+1}}}}]
\end{aligned},
\end{equation}
where $M_i$ and $M_{i+1}$ are the fan-in and fan-out of the units in the $i-$th layer, $U[x1,x2]$ denotes the uniform distribution supported by $x1$ and $x2$. ${\bm b}_h$ and ${\bm b}_g$ are the bias vectors, which can be initialized as small random values or 0. $S_h$ and $S_g$ are activation functions, such as sigmoid function, or tanh function, or rectified linear units (ReLu) first proposed in \cite{Nair2010Rectified}
\begin{equation}
\label{Eq:relu}
\begin{aligned}
  {s(a)} = max(0,a)
\end{aligned},
\end{equation}
or leaky ReLu (LReLu) proposed in \cite{maas2013rectifier}\cite{xu2015empirical}
\begin{equation}
\label{Eq:leaky_relu}
\begin{aligned}
  {s(a)} = max(0,a)-\beta\times min(0,a)
\end{aligned}.
\end{equation}
The parameter $\beta$ in Equation (\ref{Eq:leaky_relu}) represents the slope of the leak. Compared with ReLu, the LReLu can keep a small gradient even though the unit is saturated. The output $G({\bf z})$ is the generated sample, which has the same size with the real data sample.

The encoding model $E$, also called encoder, is stacked with multi-layer neural networks. Assuming an encoder with only two-layer networks, i.e., the input layer and one hidden layer, data transition from the input data $\tilde{\bf x}$ to the hidden units is called encoding, which is defined as
\begin{equation}
\label{Eq:encoding}
\begin{aligned}
  E({\tilde{\bf x}}) = {S_e}({\bm W}_e{\tilde{\bf x}}+{{\bm b}_e})
\end{aligned},
\end{equation}
where ${\bm W}_e$ is the weight matrix between the input layer and the hidden layer, which can be initialized through Equation (\ref{Eq:weight_initialization}). ${\bm b}_e$ is the bias vector initialized with small random values or 0, and $S_e$ is the activation function. The output $E({\tilde{\bf x}})$ can be considered as the feature representations of the real sample data ${\tilde{\bf x}}$ in the latent space.

The discriminative model $D$, called discriminator, is also with multi-layer network structure, i.e., the input layer, multiple hidden layers, and the output layer. It takes the combination of the sample data and its latent features as the input $\bf y$ (i.e., ${\bf y}=[G({\bf z}), {\bf z}]$ or ${\bf y}=[{\tilde{\bf x}}, E({\tilde{\bf x}})]$), and outputs $D({\bf y})\in [0,1]$ to represent the probability that $\bf y$ is from the real sample rather than the generated one. Considering a discriminator with three-layer network, the discriminative process can be denoted as in Equation (\ref{Eq:d_input_hidden}) and (\ref{Eq:d_hidden_output}).
\begin{equation}
\label{Eq:d_input_hidden}
\begin{aligned}
  {\bf h}_{d} = {S_h}({\bm W}_{h}{{\bf y}}+{{\bm b}_h})
\end{aligned},
\end{equation}
\begin{equation}
\label{Eq:d_hidden_output}
\begin{aligned}
  D({\bf y}) = {S_d}({\bm W}_{d}{{\bf h}_d}+{{\bm b}_d})
\end{aligned},
\end{equation}
where $S_h$ and $S_d$ are the activation functions, ${\bm W}_{h}$ and ${\bm W}_{d}$ are the weight matrices initialized through Equation ({\ref{Eq:weight_initialization}}), ${\bm b}_h$ and ${\bm b}_d$ are the bias vectors initialized with small random values or 0, and ${\bf h}_{d}$ represents the output of the hidden layer in $D$.

Let $p_{\bf X}$ be the distribution of the real data $\bf x$ for $\bf x\in {\Phi_{\bf X}}$ (e.g. data segments), $p_{\bf Z}$ be the distribution of the sampled data $\bf z$ in $G$ for $\bf z\in {\Phi_{\bf Z}}$. In BiGANs, we train $D$ to maximize the probability of distinguishing the real samples from the generated ones (i.e., maximizing $\log D(\tilde{\bf x},E(\tilde{\bf x}))$), train $G$ to minimize the probability of $D$ making correct distinctions (i.e., minimizing $\log (1-D(G(\bf z),{\bf z}))$, and simultaneously train $E$ to map the real data $\tilde{\bf x}$ into the latent space of $G$ (i.e., introducing $p_E ({\bf z}|{\bf x})$. Thus, the objective function of training BiGANs can be defined as \cite{DBLP:journals/corr/DonahueKD16}
\begin{equation}
\label{Eq:objective}
\begin{aligned}
  \min\limits_{G,E}\max\limits_{D}V(D,E,G)
\end{aligned},
\end{equation}
where
\begin{equation}
\label{Eq:objective_specific}
\begin{aligned}
  V(D,E,G) &= \mathbb{E}_{{\bf x}\sim p_{\bf X}}\log D(\tilde{\bf x},E(\tilde{\bf x})) \\
  &+\mathbb{E}_{{\bf z}\sim p_{\bf Z}}\log (1-D(G(\bf z),{\bf z}))
\end{aligned}.
\end{equation}

Considering the large number of parameters in BiGANs, it is mandatory to introduce an regularization technique to prevent the overfitting problem. Dropout, first proposed in \cite{Srivastava2014Dropout}, addresses this problem by introducing randomness, i.e., dropping out the units in the hidden layers with a fixed probability, such as 0.2.
The minimax objective function in Equation (\ref{Eq:objective}) can be optimized by using stochastic gradient descent (SGD) based techniques, such as adaptive subgradient (AdaGrad) method \cite{duchi2011adaptive}, root mean square prop (RMSprop) alogithm \cite{tieleman2012lecture}, adaptive moment (Adam) estimation \cite{kingma2014adam}, etc. Here, we choose Adam as the optimization algorithm, which combines the advantages of AdaGrad and RMSprop, i.e., sparse gradients, online and non-stationary settings.

In practice, the objective function in Equation (\ref{Eq:objective}) may not provide sufficient gradients for $G$ to learn well, because $D$ can clearly distinguish real sample from the generated one early in learning and this will easily lead to $\log (1-D(G(\bf z),{\bf z}))$ to saturate. Therefore, we can train $G$ by maximizing $\log (D(G(\bf z),{\bf z}))$ instead of minimizing $\log (1-D(G(\bf z),{\bf z}))$. Theoretical results in \cite{DBLP:journals/corr/DonahueKD16} show the objective function $V(D,E,G)$ achieves its global minimum value $-\log 4$ if and only if $[G({\bf z}), {\bf z}]$ and $[{\tilde{\bf x}}, E({\tilde{\bf x}})]$ have the same distribution (i.e., $p_{G({\bf z}),{\bf z}}=p_{{\tilde{\bf x}}, E({\tilde{\bf x}})}$).

The process of feature learning is training BiGANs, i.e., obtaining the optimal parameters $\theta_{D},\theta_{E},\theta_{G}$ by minimaximizing the objective function in Equation (\ref{Eq:objective}). Here, $\theta_D,\theta_E$ and $\theta_G$ denote the corresponding parameters (i.e., $\bm W$ and $\bm b$) in $D,E$ and $G$. When the network almost converges (i.e., $G \approx E^{-1}$), the features output by the encoder can be considered as the latent representations of the real data in the generator's space, which can be used for the subsequent anomaly detection task.
\subsection{Anomaly Detection}
\label{subsection: anomaly_detection}
Based on the features extracted through BiGANs for each data segment, a high-dimensional statistical index for them is calculated to indicate the data behavior. For example, at the sampling time $t_j$, the statistical index for the learned features ${\bm f}(t_j)$ is calculated as
\begin{equation}
\label{Eq:statistical_indicator}
\begin{aligned}
  \mathcal{N}_{\phi}\left( {t_j}  \right) = \sum\limits_{|{\emph f}_{r}(t_j)|> median (|{\bm f}(t_j)|)} {\phi \left( |{{{\emph f}_{r}(t_j)}}| \right)}
\end{aligned},
\end{equation}
where ${{{\emph f}_{r}(t_j)}}\in {\bm f}(t_j)$. The test function $\phi (\cdot)$ makes a linear or nonlinear mapping for the features, which can be chebyshev polynomial (CP), information entropy (IE), likelihood radio function (LRF) or wasserstein distance (WD). Detailed information about the test functions can be found in \cite{Shi2018Incipient}. $\mathcal{N}_{\phi}$ is a complex function of the extracted features, which will be further discussed in Section \ref{subsection: case_synthetic_data}.

Considering random weight initialization and dropout enforces randomness during the adversarial feature learning, the average value of the objective function in continuous $n$ iterations is calculated to judge whether terminating the training. For example, for the $i-$th iteration, the average value is calculated as
\begin{equation}
\label{Eq:obj_avg}
\begin{aligned}
 V_{\text{avg}}(D,E,G)=\frac{1}{n}\sum\limits_{l=i-n+1}^{i}V_l(D,E,G)
\end{aligned},
\end{equation}
where $V_l(D,E,G)$ is the calculated objective function value in the $l-$th ($l=i-n+1,\cdots,i$) iteration, and $V_{\text{avg}}(D,E,G)$ denotes the average value for continuous $n$ iterations. Here, the simple averaging method in ensemble learning is used. For each iteration, one network learning model is built and outputs $V_l(D,E,G)$. Thus, in continuous $n$ iterations, the average result $V_{\text{avg}}(D,E,G)$ for $n$ learning models is more accurate in judging whether to terminate the training than simply using the output $V_i(D,E,G)$ in the last iteration, because the former is more stable and reliable for reducing the error caused by randomness and the risk of network falling into local optimum. The procedure for anomaly detection based on BiGANs is summarized as in Algorithm 1.

\begin{table}[!t]
\label{Tab: algorithm1}
\centering
\footnotesize
\begin{tabular}{p{8.4cm}}   
\toprule[1.0pt]
\textbf {Algorithm 1:} The proposed BiGANs-based algorithm for anomaly detection. For $D,G,E$, the LReLu function is used as the activation function in hidden layers and the tanh function as that in output layers, and Adam is chosen as the optimization method, see Section \ref{subsection: adversarial_feature_learning} for details. $m$ denotes the number of steps applied to $D$ and $n$ is the number of iterations used to calculate $V_{\text{avg}}(D,E,G)$.  \\
\hline
\textbf{Input:} The data segment ${\{{\bf x}({t_{{N_w}+k\times N_s}})\}_{{k=1}}^{{(T-{N_w})}/{N_s}}}$, the required \\ \quad\quad\quad approximation error $\varepsilon$; \\
\textbf{Output:} The anomaly index $\mathcal{N}_{\phi}$; \\
\quad 1. \textbf{For} each data segment ${\bf x}$ \textbf{do} \\
\quad 2. \quad Normalize ${\bf x}$ into ${\tilde{\bf x}}$ according to Equation (\ref{Eq:min_max_normalize}); \\
\quad 3. \quad Initialize $\theta_D,\theta_E,\theta_G$ as illustrated in Section \ref{subsection: adversarial_feature_learning};  \\
\quad 4. \quad \textbf{For} iteration $i=1,2,3,\cdots$ \textbf{do} \\
\quad 5. \quad\quad \textbf{For} $m$ steps \textbf{do} \\
\quad 6. \quad\quad\quad Sample ${\bf z}^{(i,m)}$ from a simple latent distribution; \\
\quad 7. \quad\quad\quad Update $D,E$ by descending their gradients: \\
\quad\quad\quad\quad\quad $\nabla_{\theta_{D}}[-\log D(\tilde{\bf x},E(\tilde{\bf x}))-\log (1-D(G({\bf z}^{(i,m)}),{\bf z}^{(i,m)}))]$ \\
\quad\quad\quad\quad\quad $\nabla_{\theta_{E}}[-\log D(\tilde{\bf x},E(\tilde{\bf x}))-\log (1-D(G({\bf z}^{(i,m)}),{\bf z}^{(i,m)}))]$ \\
\quad\quad\quad\quad \textbf{End for} \\
\quad 8. \quad\quad Sample ${\bf z}^{(i)}$ from the same distribution as in step 6; \\
\quad 9. \quad\quad Update $G$ by descending its gradient: \\
\quad\quad\quad\quad $\nabla_{\theta_{G}}[-\log (D(G({\bf z}^{(i)}),{\bf z}^{(i)}))]$ \\
\quad 10. \quad\; \textbf{If} $i\ge n$ \textbf{do} \\
\quad 11. \quad\quad\quad Calculate $V_{\text{avg}}(D,E,G)$ through Equation (\ref{Eq:obj_avg}); \\
\quad 12. \quad\quad\quad \textbf{If} $|V_{\text{avg}}(D,E,G)+\log4|<\varepsilon$ \textbf{do} \\
\quad 13. \quad\quad\quad\quad $ii=i$; \\
\quad 14. \quad\quad\quad\quad Output the learned features $\{{\bf f}_l\}_{l=ii-n+1}^{ii}$ calculated \\
\quad\quad\quad\quad\quad\quad\; through Equation (\ref{Eq:encoding}); \\
\quad 15. \quad\quad\quad\quad Break; \\
\quad\quad\quad\quad\quad\; \textbf{End if} \\
\quad\quad\quad\quad\; \textbf{End if} \\
\quad\quad\quad\; \textbf{End for} \\
\quad 16. \quad Calculate $\{\mathcal{N}_{{\phi},l}\}_{l=ii-n+1}^{ii}$ through Equation (\ref{Eq:statistical_indicator}); \\
\quad 17. \quad Calculate the statistical index for $\bf x$: \\
\quad\quad\quad\; $\mathcal{N}_{\phi}=\frac{1}{n}\sum\limits_{l=ii-n+1}^{ii}\mathcal{N}_{{\phi},l}$; \\
\quad\quad\; \textbf{End for} \\
\hline
\end{tabular}
\end{table}

\section{Operational Risk Assessment Using Online Monitoring Data in Distribution Networks}
\label{section: application}
In this section, by using the online monitoring data, a new unsupervised learning approach to assess the operational risks of feeder lines in a distribution network is proposed. First, a spatio-temporal data matrix is formulated for each feeder line by using the online monitoring data, and the anomaly index $\mathcal{N}_\phi$ is calculated as illustrated in Section \ref{section: theory}. Then, by combining the confidence level $1-\alpha$ for the population mean of the standardized  $\mathcal{N}_\phi$, the operational risks are classified into different levels with clear criterion defined. The specific steps of the proposed approach are given and analyzed.
\subsection{Formulation of Online Monitoring Data as Spatio-Temporal Matrices}
\label{subsection: data_matrix_formulation}
\begin{figure}[!t]
\centerline{
\includegraphics[width=2.5in]{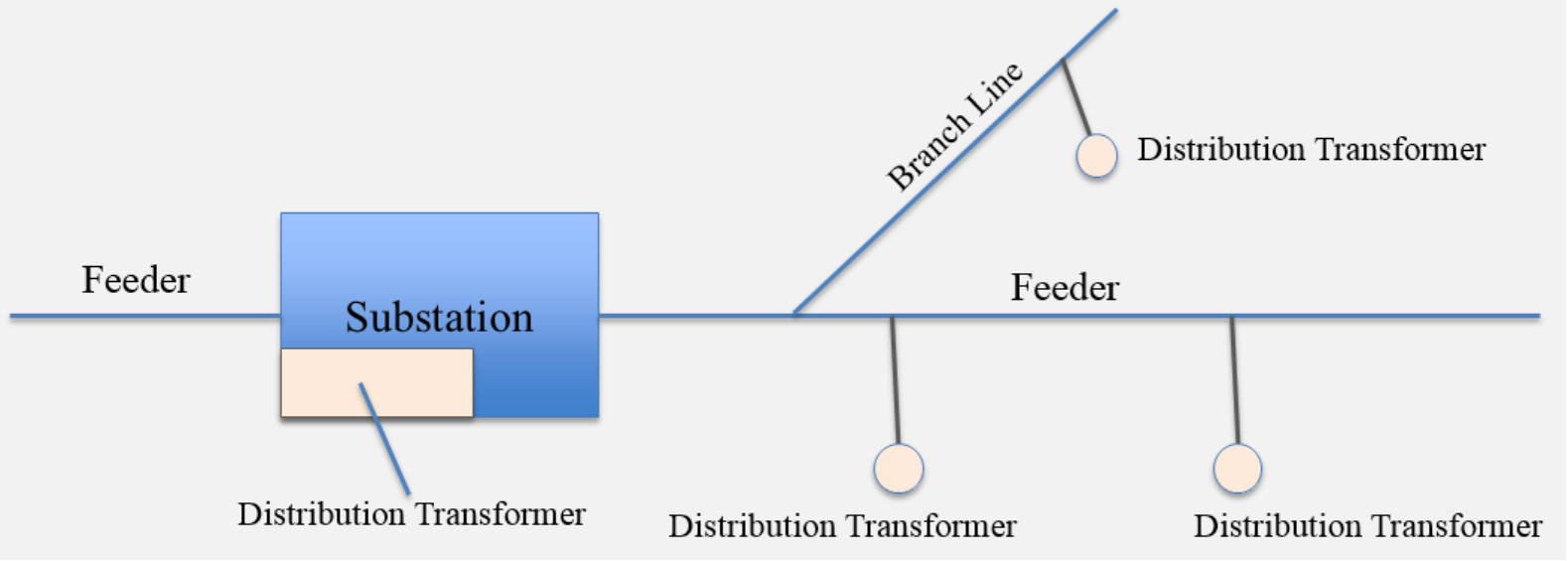}
}
\caption{Circuitry topology diagram of partial distribution network.}
\label{fig:circuitry_topology}
\end{figure}
As illustrated in Figure \ref{fig:circuitry_topology}, one feeder line in the partial distribution network consists of branch lines and substations with distribution transformers. On the low voltage side of each distribution transformer, one online monitoring sensor is installed, through which we can obtain multiple measurements, such as three-phase voltages (i.e.,$u_a,u_b,u_c$). Here, we choose $u_a,u_b,u_c$ at the sampling time $t_j$  to formulate a data vector ${{\bf{d}}(t_j)} = {\left[ {{{u}}_{aj}^{\left( 1 \right)},{{u}}_{bj}^{\left( 1 \right)},{{u}}_{cj}^{\left( 1 \right)}, \cdots , {{u}}_{aj}^{\left( m \right)},{{u}}_{bj}^{\left( m \right)},{{u}}_{cj}^{\left( m \right)}} \right]^H}$, where $m$ denotes the number of sensors installed on the feeder. Let $P=3m$, for a series of time $T$, we can obtain a spatio-temporal data matrix ${\bf D}=[{\bf d}(t_1),{\bf d}(t_2),\cdots,{\bf d}(t_T)]\in{\mathbb{R}^{P\times T}}$. It is noted that, by stacking the voltage measurements together, the formulated spatio-temporal data matrix contains rich information on the operating state of the feeder line.
\subsection{Operational Risk Classification in Distribution Networks}
\label{subsection: status_classification}
The anomaly detection result can indicate the operating states of the feeder lines in distribution networks. Here, it is used as the basis for assessing the operational risks. For a series of time $T'$, the anomaly index $\mathcal{N}_\phi$ for each data segment is calculated, size of which is $\frac{T'}{N_s}$. We first standardize $\mathcal{N}_\phi$ by
\begin{equation}
\label{Eq:indicator_standard}
\begin{aligned}
  {\hat{\mathcal{N}_{\phi}}} = \frac{{\mathcal{N}_{\phi}}-\mu ({\mathcal{N}_{\phi}})}{\sigma ({\mathcal{N}_{\phi}})}
\end{aligned},
\end{equation}
where $\mu ({\mathcal{N}_{\phi}})$ and $\sigma ({\mathcal{N}_{\phi}})$ are the sample mean and sample standard deviation of $ {\mathcal{N}_{\phi}}$ in a series of time $T'$.

Considering the sample size $\frac{T'}{N_s}$ is sometimes small, here, $\hat{\mathcal{N}_\phi}$ is assumed to follow a student's t-distribution with $\frac{T'}{N_s}-1$ degrees of freedom, i.e., $\hat{\mathcal{N}_\phi}\sim t(\frac{T'}{N_s}-1)$. According to the central limit theorem, the confidence level $1-\alpha$ for the population mean $\mu$ of $\hat{\mathcal{N}_\phi}$ is defined as
\begin{equation}
\label{Eq:confidence_level}
\begin{aligned}
  1-\alpha = P\{{\mu(\hat{\mathcal{N}_\phi})-t_{\frac{\alpha}{2}}{\frac{\sigma(\hat{\mathcal{N}_\phi})}{\sqrt{\frac{T'}{N_s}}}}}<\mu<{\mu(\hat{\mathcal{N}_\phi})+t_{\frac{\alpha}{2}}{\frac{\sigma(\hat{\mathcal{N}_\phi})}{\sqrt{\frac{T'}{N_s}}}}}\}
\end{aligned},
\end{equation}
where $\mu(\hat{\mathcal{N}_\phi})$ and $\sigma(\hat{\mathcal{N}_\phi})$ are the sample mean and sample standard deviation of $\hat{\mathcal{N}_\phi}$ with $\mu(\hat{\mathcal{N}_\phi})=0$ and $\sigma(\hat{\mathcal{N}_\phi})=1$, $t_{\frac{\alpha}{2}}$ is the upper $\frac{\alpha}{2}$ critical value for the t distribution with $\frac{T'}{N_s}-1$ degrees of freedom, and $P\{\cdot\}$ is the probability operator. Thus, the confidence interval of level $1-\alpha$ is simplified as $[-t_{\frac{\alpha}{2}}{\frac{1}{\sqrt{\frac{T'}{N_s}}}},t_{\frac{\alpha}{2}}{\frac{1}{\sqrt{\frac{T'}{N_s}}}}]$, and the p value for the interval critical values is equal to $\frac{\alpha}{2}$. For a given $\hat{\mathcal{N}_\phi}$, the corresponding p value can be obtained by the t distribution table. For example, let ${\hat{\mathcal{N}_{\phi}}}=2.650$ and $\frac{T'}{N_s}-1=13$, then the p value is $1\%$.

To further quantify the operational risks of feeder lines in distribution networks, we classify the operational risk levels into emergency, high risk, preventive and normal according to the defined intervals of the confidence level $1-\alpha$ for the population mean of $\hat{\mathcal{N}_\phi}$, which is shown in Table \ref{Tab:state_classification}. Thus, for a calculated $\hat{\mathcal{N}_\phi}$, we can judge the operational risk level by comparing the p value with the corresponding interval of $\frac{\alpha}{2}$: the smaller the p value, the higher the risk level.
\begin{table}[htbp]
\caption{The Classification Results of Operational Risk Levels.}
\label{Tab:state_classification}
\centering


\footnotesize
\begin{tabular}{cc}   
\toprule[1.0pt]
\textbf{Operational risk level}& \textbf{Confidence level ($1-\alpha$)}\\
\hline
Emergency & $>97.5\%$ \\
High risk & $95\% <\cdots\le 97.5\%$ \\
Preventive & $90\% <\cdots\le 95\%$ \\
Normal & $\le 90\%$ \\
\hline
\end{tabular}
\end{table}

Here, ``emergency'' means a feeder line operates in abnormal state and serious faults may happen at any time. If one feeder is diagnosed as in emergency state, it will be further analyzed. ``High risk'' denotes a feeder line is of high risk in suffering from faults, which deserves special attention. ``Preventive'' means a feeder line operates in normal state, but it is not safe and should be watched for a period of time. ``Normal'' denotes a feeder line is in healthy state. By using Table \ref{Tab:state_classification}, the operational risks of feeders are quantified, which offer references for operators to make safety assessments.
\subsection{The Operational Risk Assessment Approach in Distribution Networks}
\label{subsection: state_assessment}
Based on the research above, an unsupervised risk assessment approach in distribution networks is proposed. The steps of the approach are shown as follows.
\begin{table}[htbp]
\label{Tab:algorithm2}
\centering
\footnotesize
\begin{tabular}{p{8.4cm}}   
\toprule[1.0pt]
\textbf {Steps of the operational risk assessment in distribution networks}\\
\hline
1. For each feeder, a spatio-temporal data matrix ${\bf D}\in{\mathbb{R}^{P\times T}}$ is \\
\quad formulated as illustrated in Section \ref{subsection: data_matrix_formulation}.  \\
2. Partition $\bf D$ into a series of data segments with a $P\times N_w$ window \\
\quad moving on it at a step size $N_s$. \\
3. For the data segment at the sampling time $t_j$, \\
\quad 3a) Reshape it into a column vector ${\bf x}(t_j)$; \\
\quad 3b) Normalize ${\bf x}(t_j)$ into ${\tilde{\bf x}}(t_j)$ according to Equation (\ref{Eq:min_max_normalize}); \\
\quad 3c) calculate the anomaly detection index $\mathcal{N}_{\phi}(t_j)$, see Algorithm 1 \\
\quad\quad\; for details. \\
4. Draw $\mathcal{N}_{\phi}-t$ curve for each feeder in a series of time $T'$. \\
5. Calculate the p value for each data point of $\hat{\mathcal{N}_\phi}$ in Equation (\ref{Eq:indicator_standard}). \\
6. Assess the feeder's operational risk level by comparing the calculated \\
\quad p value with the interval of $\frac{\alpha}{2}$ defined in Table \ref{Tab:state_classification}. \\
\hline
\end{tabular}
\end{table}

The operational risk assessment approach proposed is driven by the online monitoring data and based on adversarial feature learning theory. Step 1 is conducted for the formulation of a spatio-temporal data matrix for each feeder. In Step 2, the data matrix is partitioned into a series of data segments by using a moving window method. Step 3 is the adversarial feature learning process for each data segment, in which no assumptions or simplifications are made for the underlying structure of the real data. Step 4$\sim$6 are conducted for the operational risk assessment based on the central limit theorem. The proposed approach is practical for online analysis when the last sampling time is considered as the current time.

\section{Case Studies}
\label{section: case}
In this section, we validate the effectiveness of the proposed approach and compare it with other existing approaches. Six cases in different scenarios are designed. The first three cases, using the synthetic data generated from IEEE 118-bus test system, test the performances of the proposed approach with different parameter settings, which offer parameter selection guidelines for analyzing the real data. The last three cases, using the real-world online monitoring data, validate the proposed approach and compare it with other existing approaches.
\subsection{Case Study with Synthetic Data}
\label{subsection: case_synthetic_data}
The synthetic data was sampled from the simulation results of the IEEE 118-bus test system \cite{5491276}. In the simulations, a sudden change of the active load at one bus was considered as an anomaly signal and a little white noise was introduced to represent random fluctuations.

1) Case Study on $z-$sampling Distribution: In BiGANs, $z$ represents the input data of the generative model $G$, which is sampled from a simple distribution, such as uniform distribution, gaussian distribution, exponential distribution, etc. In this case, we will explore whether $z-$sampling distribution affects the proposed approach's performance. The synthetic data set contained 118 voltage measurements for sampling 500 times. An assumed step signal was set for bus 20 during $t_s=251\sim 255$ and others stayed unchanged, which was shown in Table \ref{Tab: Case1}.
\begin{table}[!t]
\caption{An Assumed Signal for Active Load of Bus 20 in Case 1.}
\label{Tab: Case1}
\centering
\footnotesize
\begin{tabular}{clc} 
\toprule[1.0pt]
\textbf{Bus} & \textbf{Sampling Time}& \textbf{Active Load(MW)}\\
\hline
\multirow{3}*{20} & $t_s=1\sim 250$ & 20 \\
~&$t_s=251\sim 255$ & 120 \\
~&$t_s=256\sim 500$ & 20 \\
\hline
Others & $t_s=1\sim 500$ & Unchanged \\
\hline
\end{tabular}
\end{table}
The other involved parameters were set as follows:\\
-- The moving window's size $P\times N_w$: $118\times 10$; \\
-- The moving step size $N_s$: $10$; \\
-- The number of layers for $D$/$E$/$G$: $5$; \\
-- The number of neurons in each hidden layer of $D$:

$768,320,256$; \\
-- The number of neurons in each hidden layer of $E$:

$768,320,256$; \\
-- The number of neurons in each hidden layer of $G$:

$256,320,768$; \\
-- The feature size: 64; \\
-- The number of steps $m$ applied to $D$: 1; \\
-- The number of iterations $n$ to calculate $V_{\text{avg}}(D,E,G)$: 10; \\
-- The initial learning rate $\eta$: 0.0002; \\
-- The slope of the leak $\beta$ in LReLu: 0.2; \\
-- The dropout coefficient: 0.1; \\
-- The required approximation error $\varepsilon$: 0.0001; \\
-- The test function $\phi(\lambda)$: $-\lambda$ln$(\lambda)$. \\

\begin{figure}[!t]
\centerline{
\includegraphics[width=2.5in]{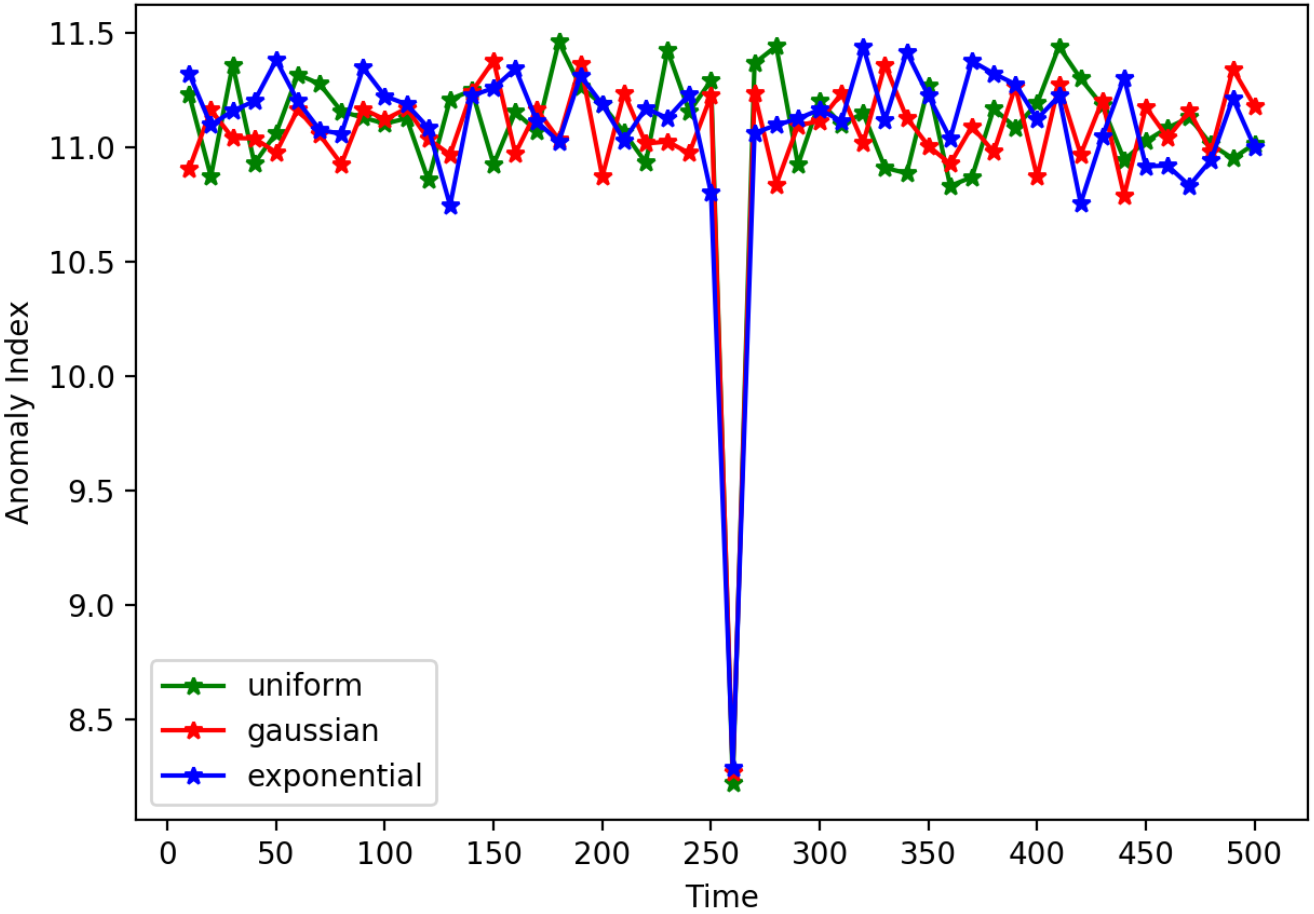}}
\caption{The anomaly detection results with different $z-$sampling distribution.}
\label{fig:distribution}
\end{figure}
For exploring the effect of $z-$sampling distribution on the performance of the proposed approach, the statistical indices with $z$ sampled from uniform distribution (i.e., $z\sim U(0,1)$), gaussian distribution (i.e., $z\sim N(0,1)$) and exponential distribution (i.e., $z\sim E(1)$) were respectively calculated and the corresponding $\mathcal{N}_\phi-t$ curves were plotted in Figure \ref{fig:distribution}. It can be observed that the assumed anomaly signal can be detected when $z$ is sampled from any distribution. Meanwhile, for the $\mathcal{N}_\phi-t$ curves, the p values of $\hat{\mathcal{N}_\phi}$ corresponding to the anomaly point were calculated, results of which were $0.0005\%$, $0.0005\%$, $0.0005\%$, respectively. It can be concluded that the detection performance of the proposed approach is almost not affected by the assumption of $z-$sampling distribution.

\begin{figure}[!t]
\centerline{
\includegraphics[width=2.5in]{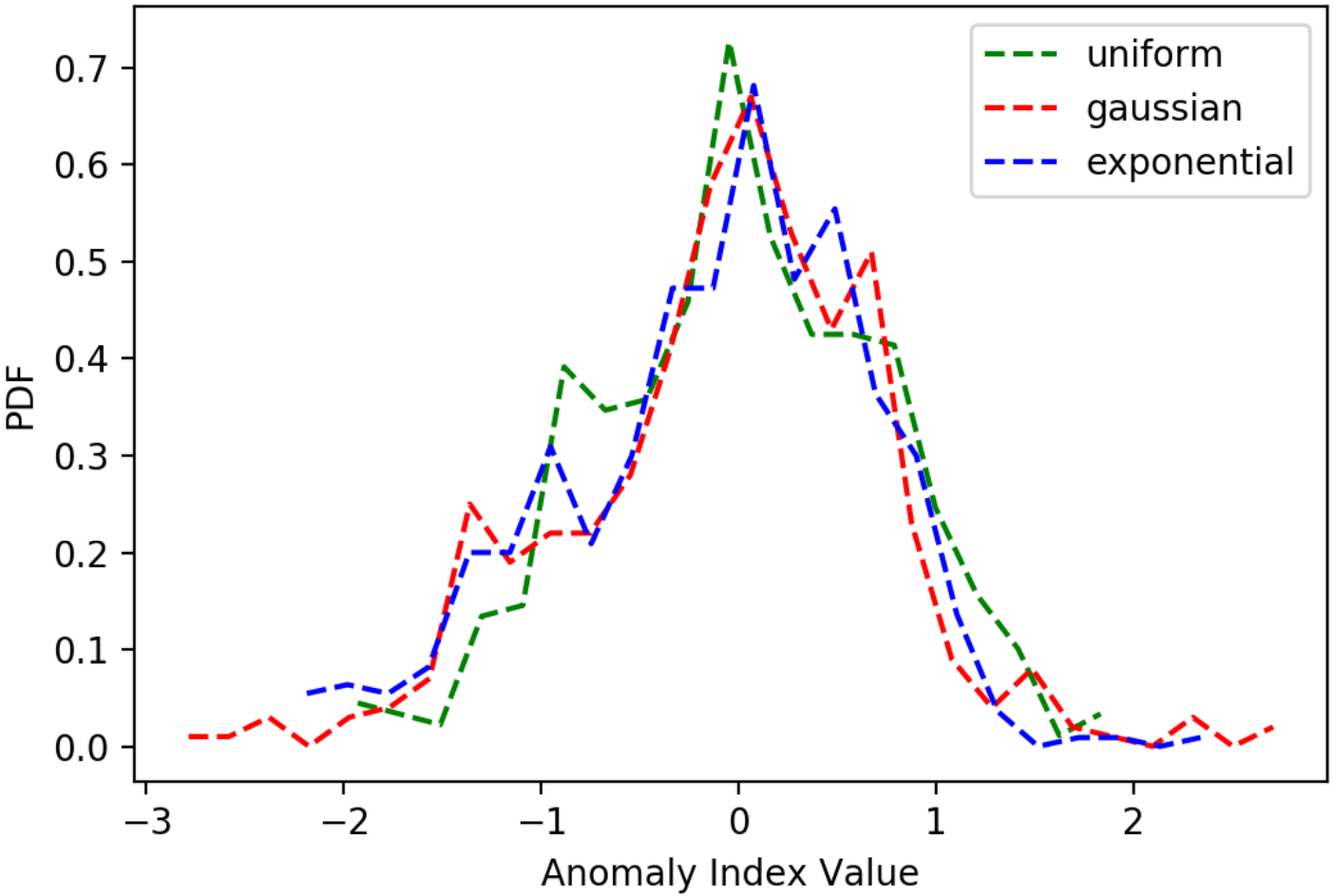}}
\caption{The shape of sampling distribution of the standardized anomaly indices with different $z-$sampling distribution.}
\label{fig:distribution_pdf}
\end{figure}
In the experiment, the synthetic data set were partitioned into $50$ data segments and $\mathcal{N}_{{\phi},l} (l=1,\cdots,n=10)$ were generated for each data segment. In order to explore the shape of the sampling distribution of the anomaly index corresponding to different $z-$sampling distribution, the probability density function (PDF) curves of $\hat{\mathcal{N}_\phi}$ with outliers (the values corresponding to the anomaly point) dropped are plotted in Figure \ref{fig:distribution_pdf}. It can be observed that the sampling distribution of $\hat{\mathcal{N}_\phi}$ is approximately normal when the degrees of freedom are large, regardless of $z-$sampling distribution. It validates our assumption in Section \ref{subsection: status_classification} that $\hat{\mathcal{N}_\phi}$ follows a t distribution.

\begin{figure}[!t]
\centerline{
\includegraphics[width=2.5in]{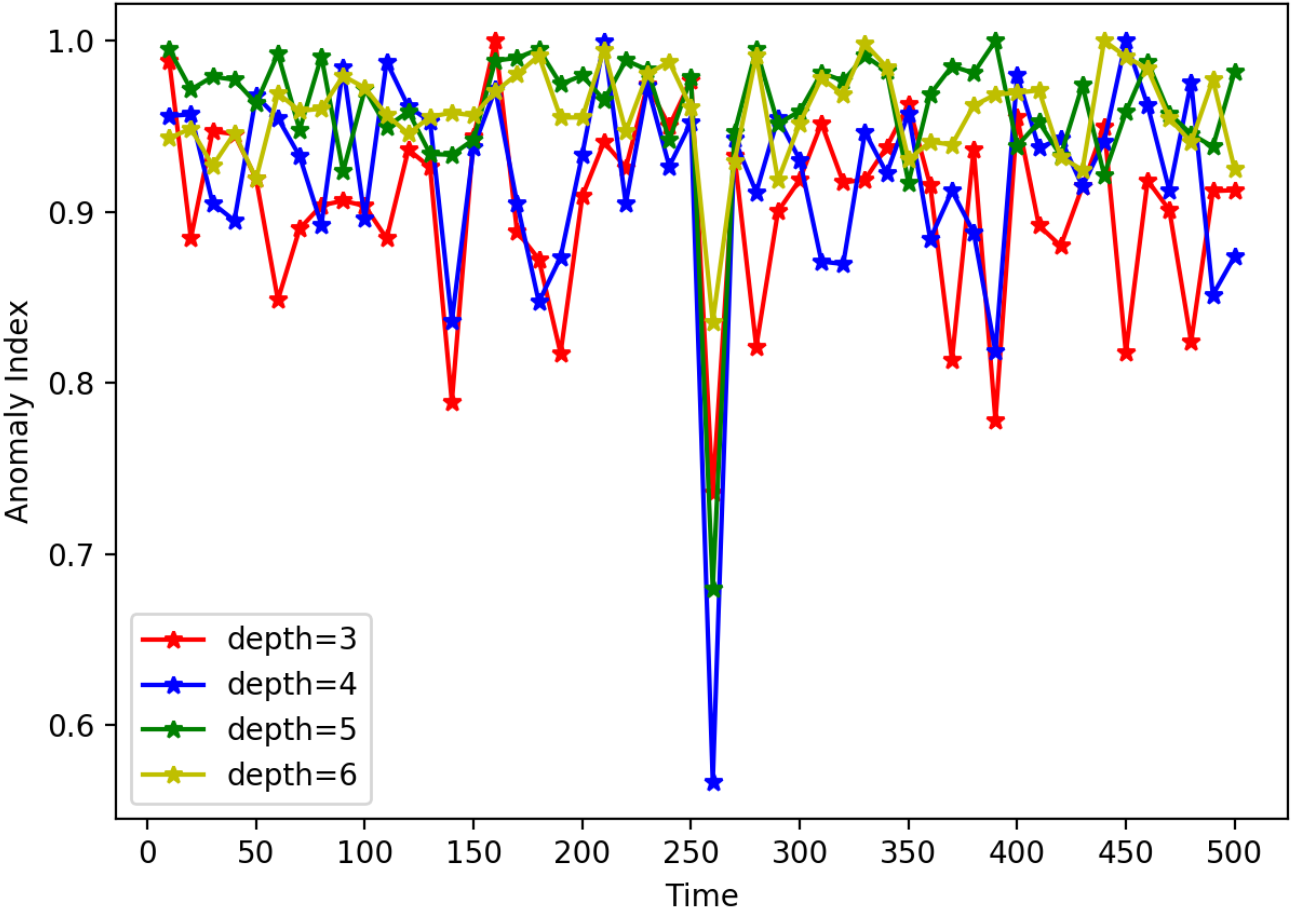}
}
\caption{The anomaly detection results with different model depth.}
\label{fig:model_depth}
\end{figure}

2) Case Study on Model Depth: Since $D, E$ and $G$ in BiGANs are composed of multi-layer network, in this case, we will explore how the model depth (i.e., the number of layers in $D$/$E$/$G$) affects the proposed approach's performance. The generated data set in Case 1) was used in this case, and $z$ was sampled from standard gaussian distribution, i.e., $z\sim N(0,1)$. The other involved parameters were set the same as in Case 1). For illustrating the effect of model depth on the performance of the proposed approach, the anomaly indices corresponding to different model depth were calculated and normalized into $[0,1]$, which was shown in Figure \ref{fig:model_depth}.

\begin{figure}[!t]
\centering
\begin{minipage}{4.1cm}
\centerline{
\includegraphics[width=.18\textheight,height=.65\textwidth]{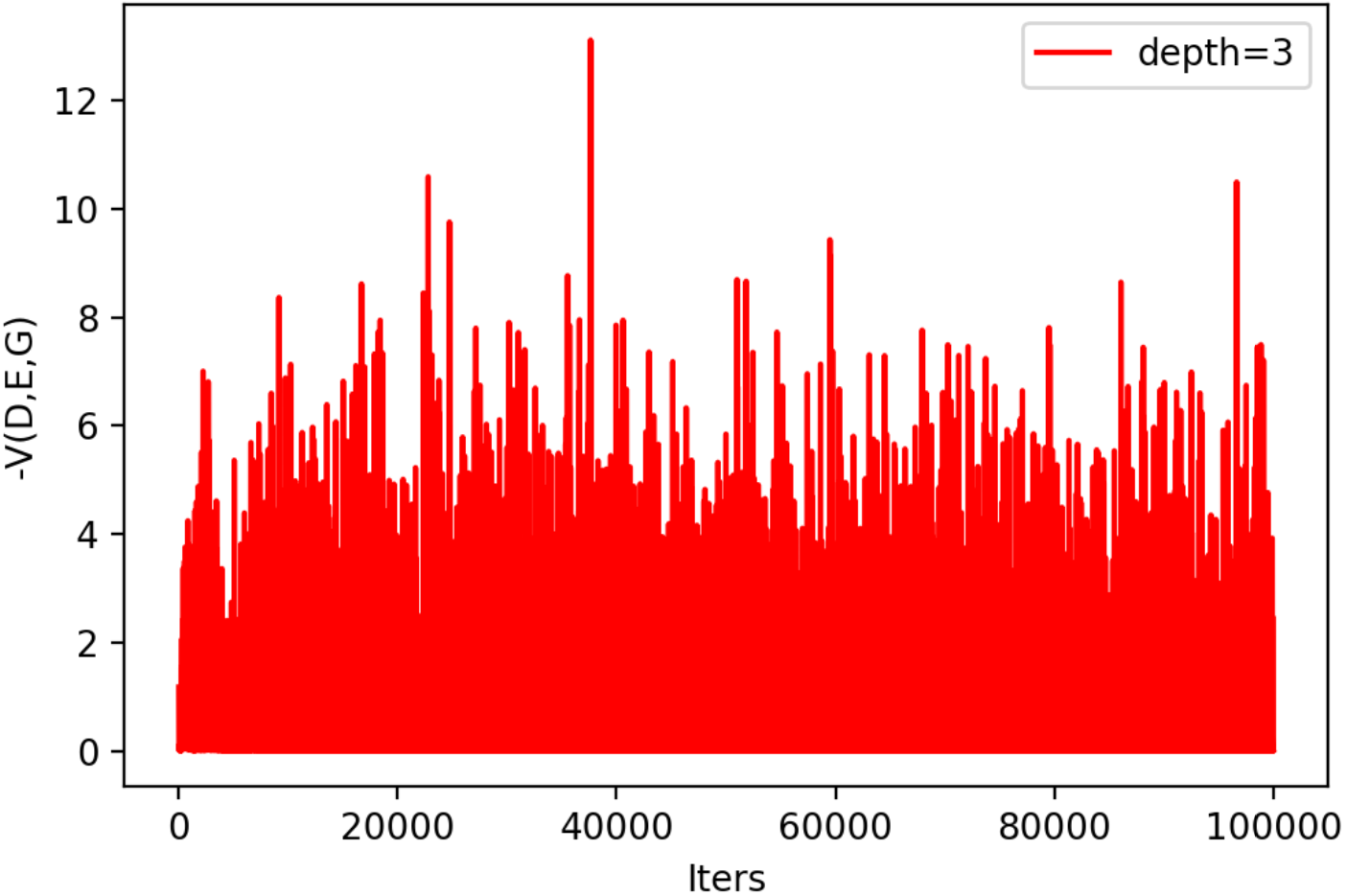}
}
\parbox{5cm}{\small \hspace{1.2cm}(a) $depth=3$}
\end{minipage}
\hspace{0.2cm}
\begin{minipage}{4.1cm}
\centerline{
\includegraphics[width=.18\textheight,height=.65\textwidth]{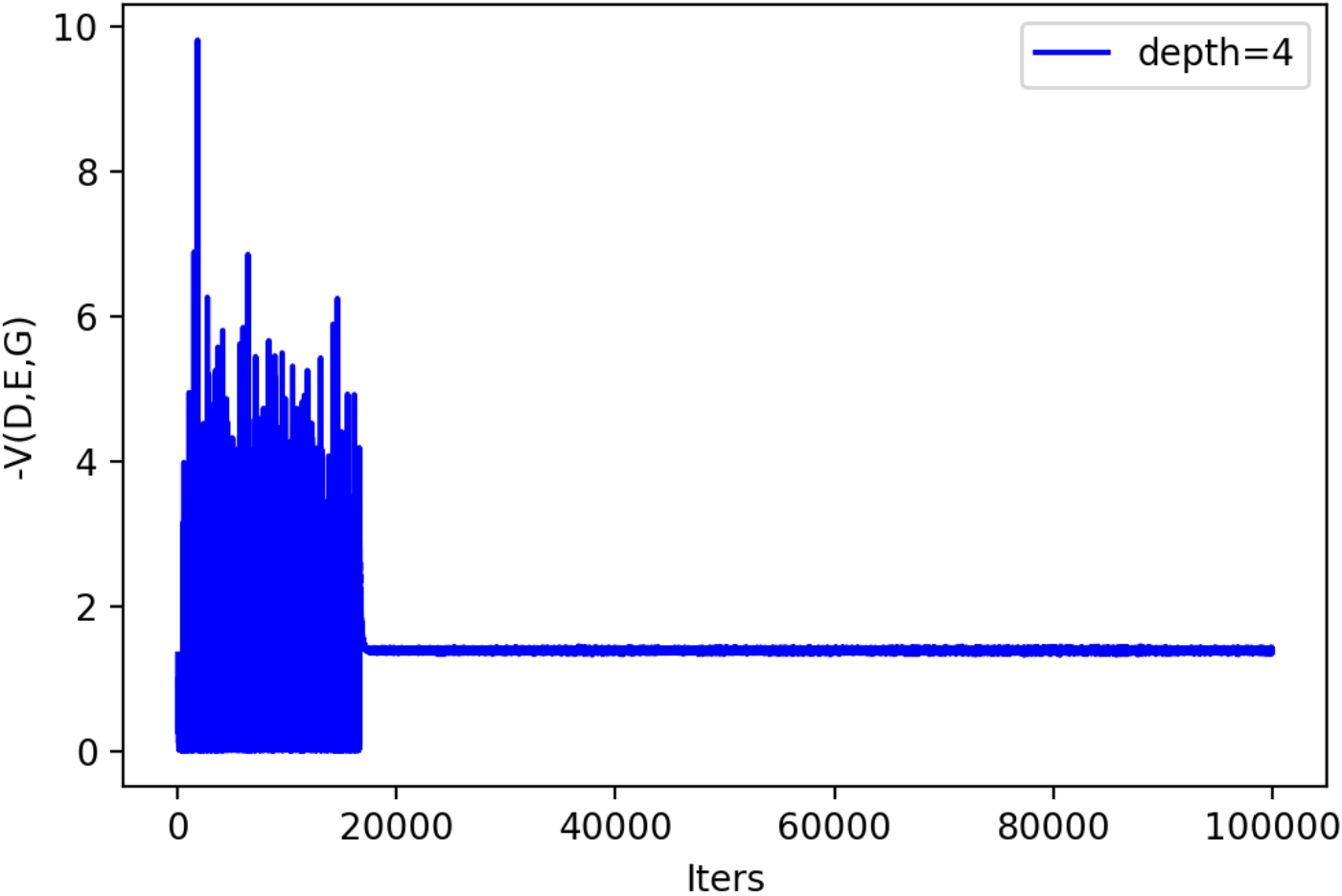}
}
\parbox{5cm}{\small \hspace{1.2cm}(b) $depth=4$}
\end{minipage}
\hspace{0.2cm}
\begin{minipage}{4.1cm}
\centerline{
\includegraphics[width=.18\textheight,height=.65\textwidth]{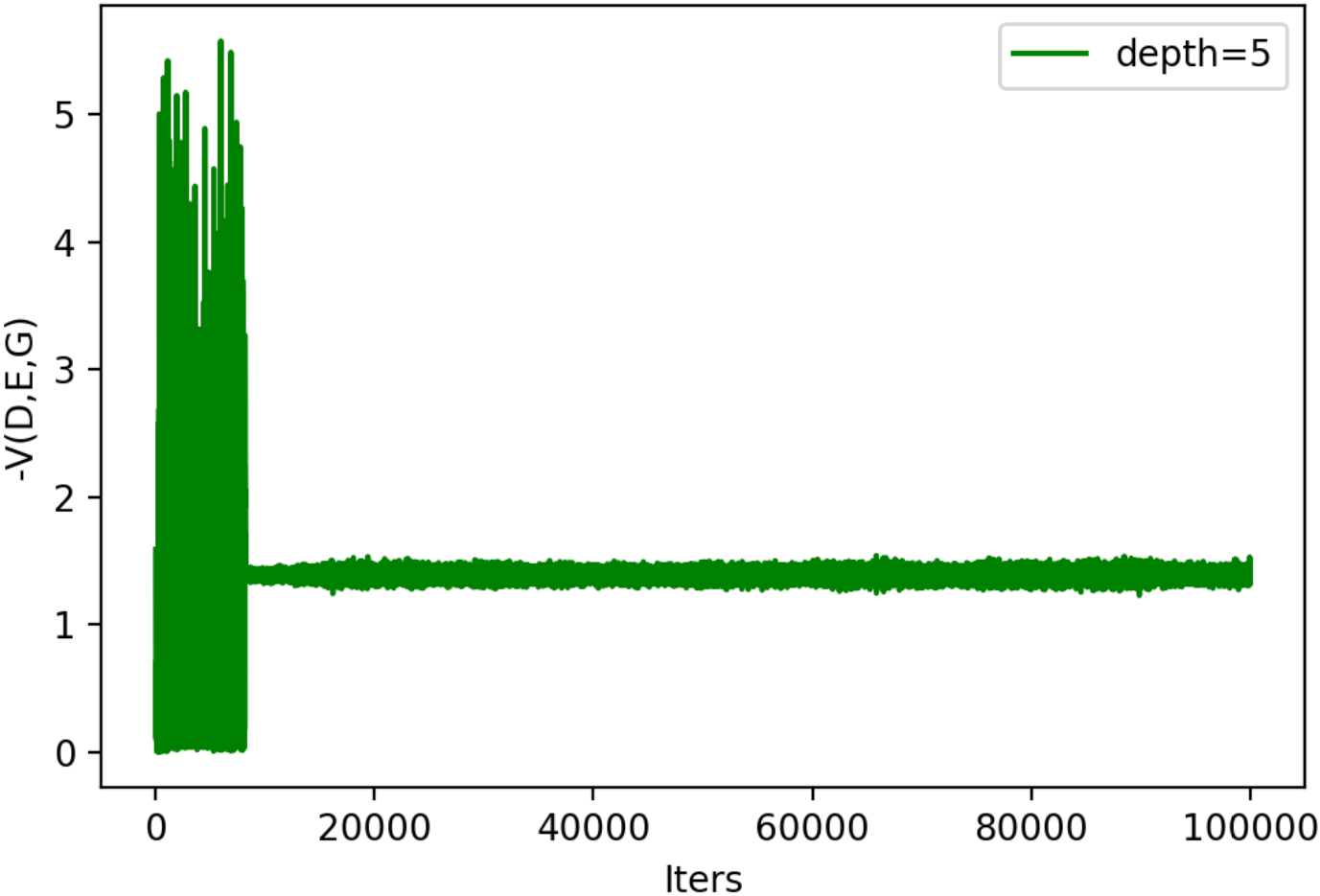}
}
\parbox{5cm}{\small \hspace{1.2cm}(c) $depth=5$}
\end{minipage}
\hspace{0.2cm}
\begin{minipage}{4.1cm}
\centerline{
\includegraphics[width=.18\textheight,height=.65\textwidth]{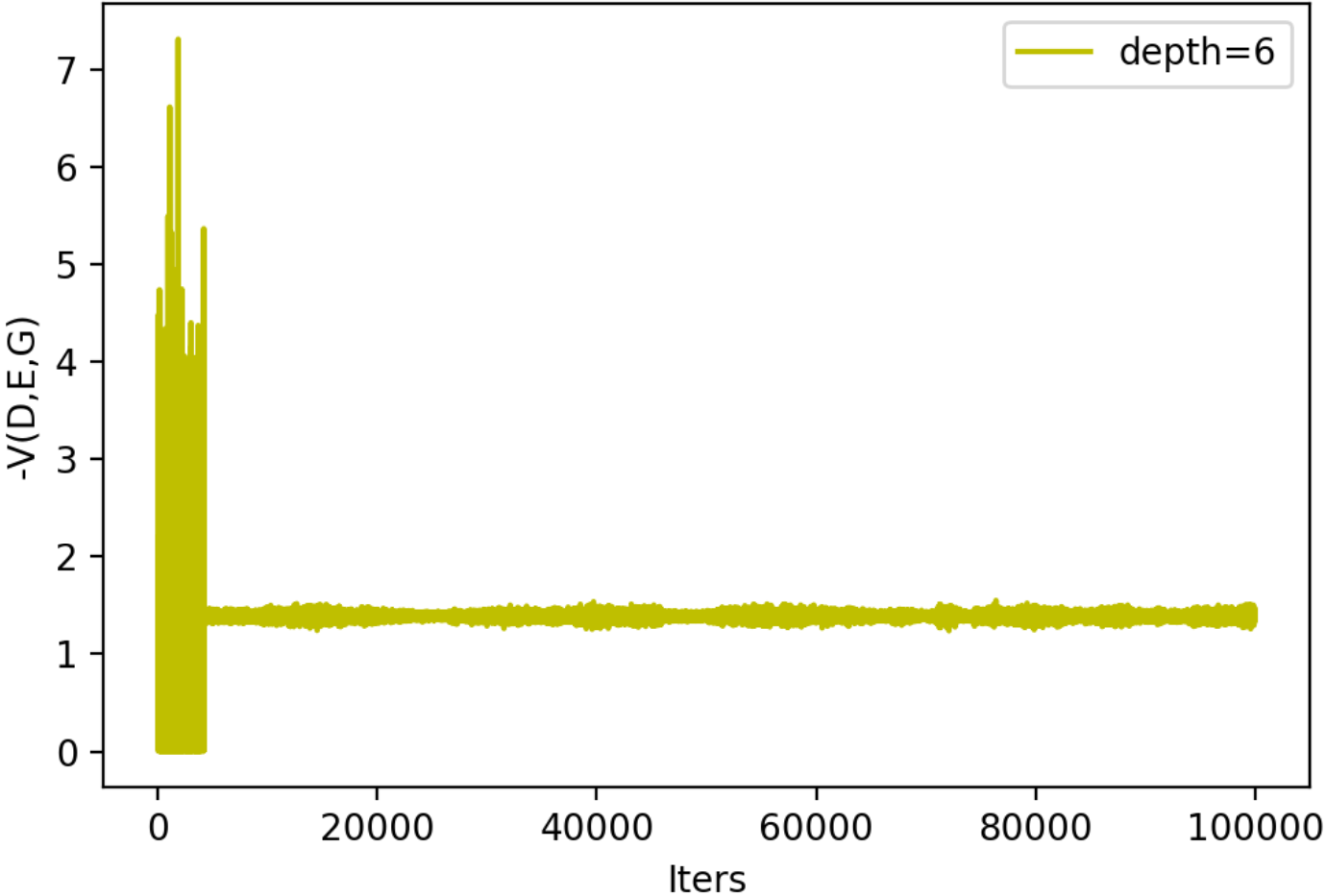}
}
\parbox{5cm}{\small \hspace{1.2cm}(d) $depth=6$}
\end{minipage}
\caption{The convergence rate of training BiGANs with different model depth. The maximum number of iterations are set to be $100000$. It can be observed that, with the increase of model depth, the number of iterations to converge for training BiGANs become less.}
\label{fig:convergence_speed}
\end{figure}

It can be observed that the assumed anomaly signal can be detected for different model depth (i.e., $depth=3,4,5,6$). Meanwhile, for the  $\mathcal{N}_\phi-t$ curves, the p values of $\hat{\mathcal{N}_\phi}$ corresponding to the anomaly point were calculated, results of which were $0.1815\%$, $0.0005\%$, $0.0005\%$, $0.0030\%$, respectively. It shows that the best anomaly detection performance is achieved when the model depth is $4$ or $5$. Furthermore, the effect of model depth on the convergence rate in training BiGANs is illustrated in Figure \ref{fig:convergence_speed}. Considering the performance and efficiency comprehensively, the model depth is set as $5$ in the subsequent experiments.

\begin{figure}[!t]
\centerline{
\includegraphics[width=2.5in]{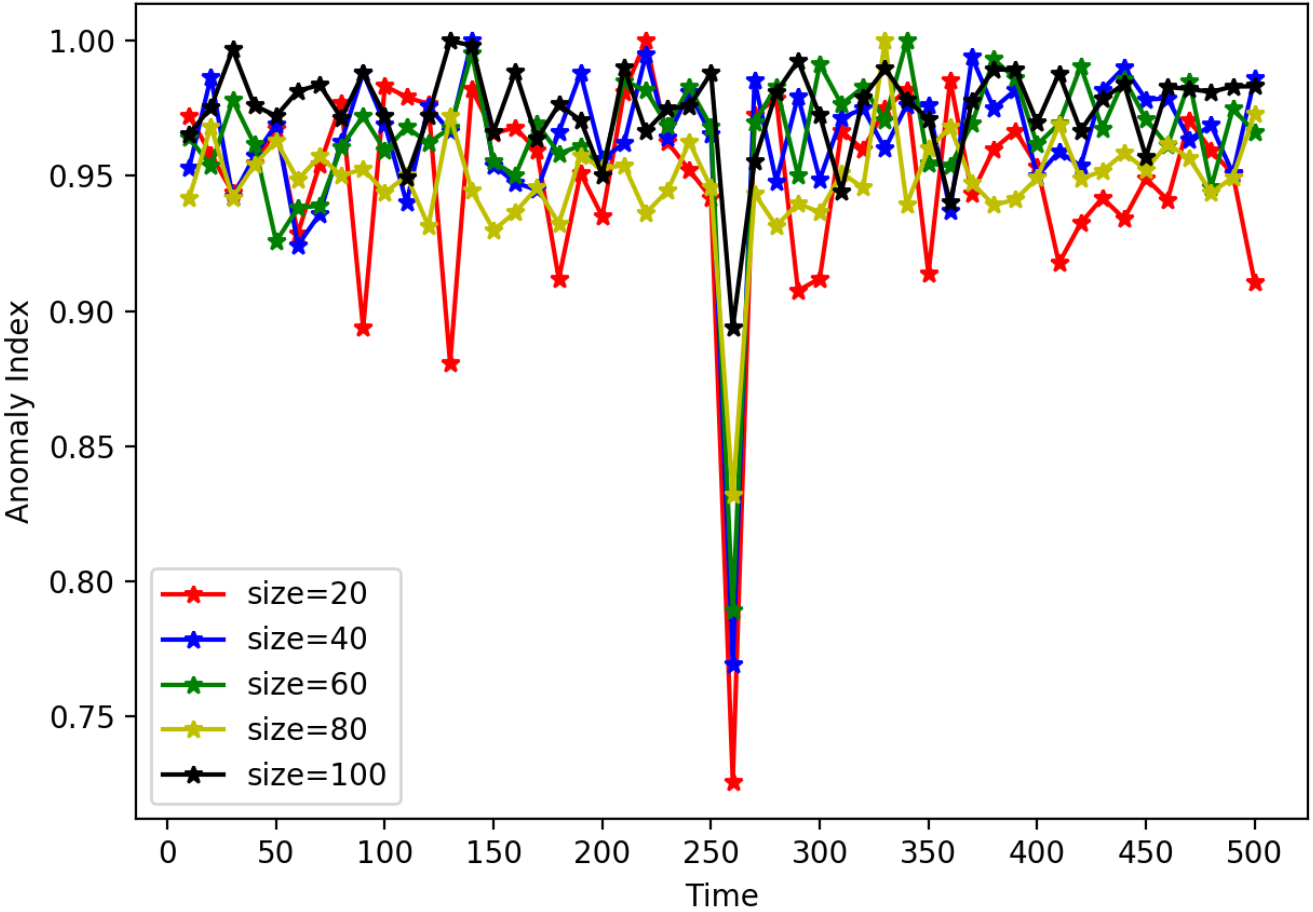}
}
\caption{The anomaly detection results corresponding to different feature size.}
\label{fig:feature_size}
\end{figure}

3) Case Study on Feature Size: In this case, the effect of feature size on the performance of the proposed approach is explored. The generated data set in case 1) was used in this case, and the model depth was set to be $5$. The other involved parameters were set the same as in Case 1). The anomaly indices corresponding to different feature size were calculated and normalized into $[0,1]$, which was shown in Figure \ref{fig:feature_size}.

It can be observed that the assumed anomaly signal can be detected for different feature size (i.e., $size=20,40,60,80,100$). For the $\mathcal{N}_\phi-t$ curve, the p values of $\hat{\mathcal{N}_\phi}$ corresponding to the anomaly point were calculated and the results were $0.0010\%$, $0.0005\%$, $0.0005\%$, $0.0005\%$, $0.0115\%$, respectively. It can be concluded that: 1) when the feature size is small (such as 20 or 40), the proposed approach is sensitive to the anomaly signal, but it is vulnerable to random fluctuations; 2) with the increase of feature size, the proposed approach becomes less sensitive to the anomaly signal and more robust to random fluctuations. In the experiment, we note that large feature size will lead to a slow convergence rate in training BiGANs. Therefore, a moderate feature size is often selected empirically.
\subsection{Case Study on Real-World Online Monitoring Data}
\label{subsection: case_real_data}
In this section, the online monitoring data obtained from a distribution network in Hangzhou city of China is used to validate the proposed approach. The distribution network contains $200$ feeder lines with $8000$ distribution transformers. The online monitoring data were sampled every 15 minutes. Anomaly time and type for each feeder line were recorded during the operation. In the following cases, three-phase voltages were chosen as the measurement variables to formulate the data matrices. Voltage violation and disturbance were considered as the risk items.

\begin{figure}[!t]
\centerline{
\includegraphics[width=2.5in]{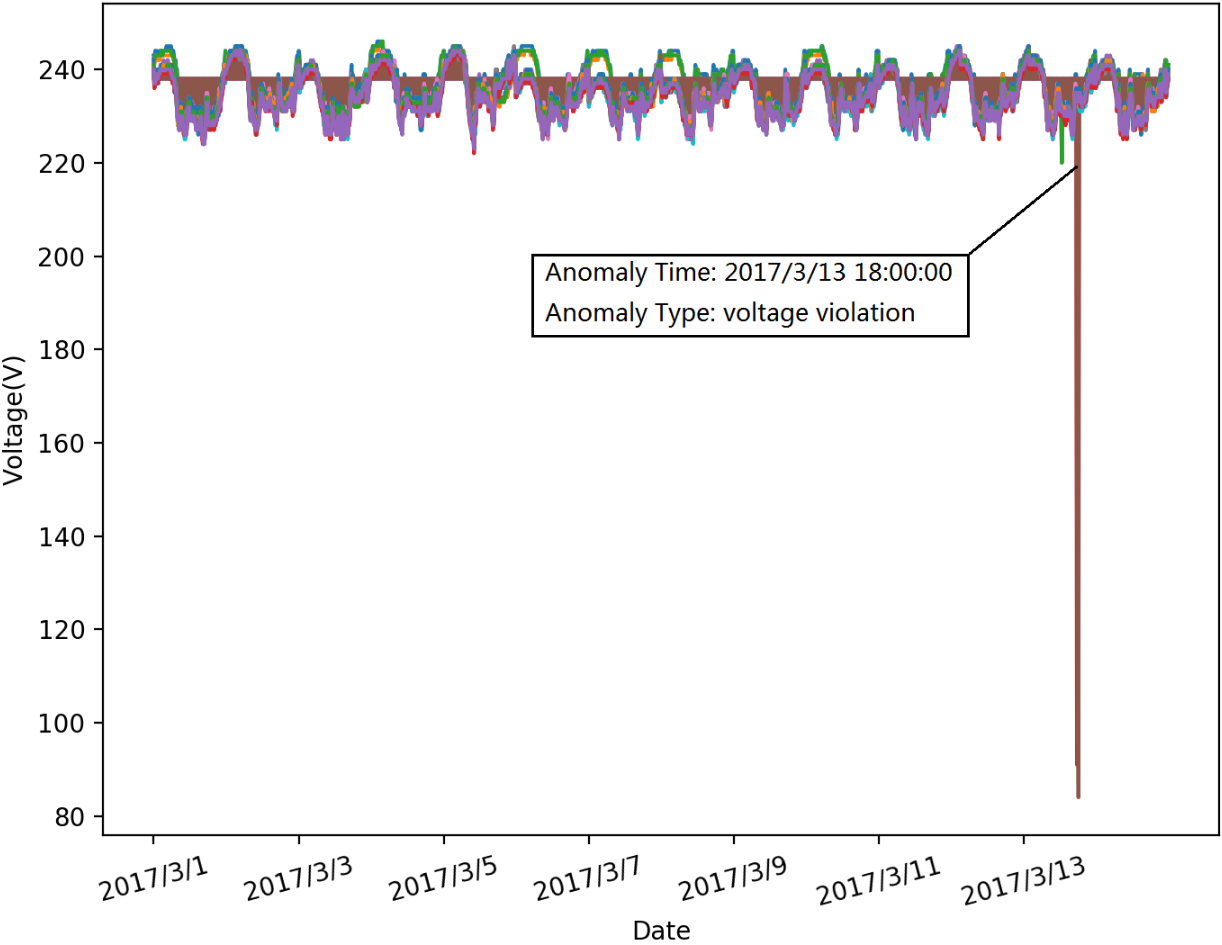}
}
\caption{The real-world online monitoring data with anomaly time and type labelled. The anomaly time is 2017/3/13 18:00:00 and the anomaly type is voltage violation.}
\label{fig:Case_voltage_violation_data}
\end{figure}

1) Case Study on Voltage Violation: Voltage violation is an common anomaly type in distribution networks, which increases the operational risks of the networks. It contains two aspects, i.e., exceeding the upper limit or the lower limit. In this case, we assess the operational risk of one feeder line suffering from voltage violation to validate the proposed approach. The feeder, with branch lines and substations, contained $15$  distribution transformers in total. The online monitoring data were sampled from 2017/3/1 00:00:00 to 2017/3/14 23:45:00, thus a $45\times 1344$ data matrix was formulated. The data with anomaly time and type labelled are shown in Figure \ref{fig:Case_voltage_violation_data}. The involved parameters are set as follows: \\
-- the moving window's size $P\times N_w$: $45\times 96$; \\
-- the moving step size $N_s$: $96$; \\
-- the model depth: $5$; \\
-- the number of neurons in each hidden layer of $D$:

$1660,960,320$; \\
-- the number of neurons in each hidden layer of $E$:

$1660,960,320$; \\
-- the number of neurons in each hidden layer of $G$:

$320,960,1660$; \\
-- the feature size: $64$; \\
-- the number of steps $m$ applied to $D$: $1$; \\
-- the number of iterations $n$ to calculate $V_{\text{avg}}(D,E,G)$: $5$; \\
-- the initial learning rate $\eta$: $0.0001$; \\
-- the slope of the leak $\beta$ in LReLu: $0.2$; \\
-- the dropout coefficient: $0.2$; \\
-- the required approximation error $\varepsilon$: $0.0001$; \\
-- the test function $\phi (\lambda)$: $-\lambda$ln$(\lambda)$. \\
\begin{figure}[!t]
\centerline{
\includegraphics[width=2.5in]{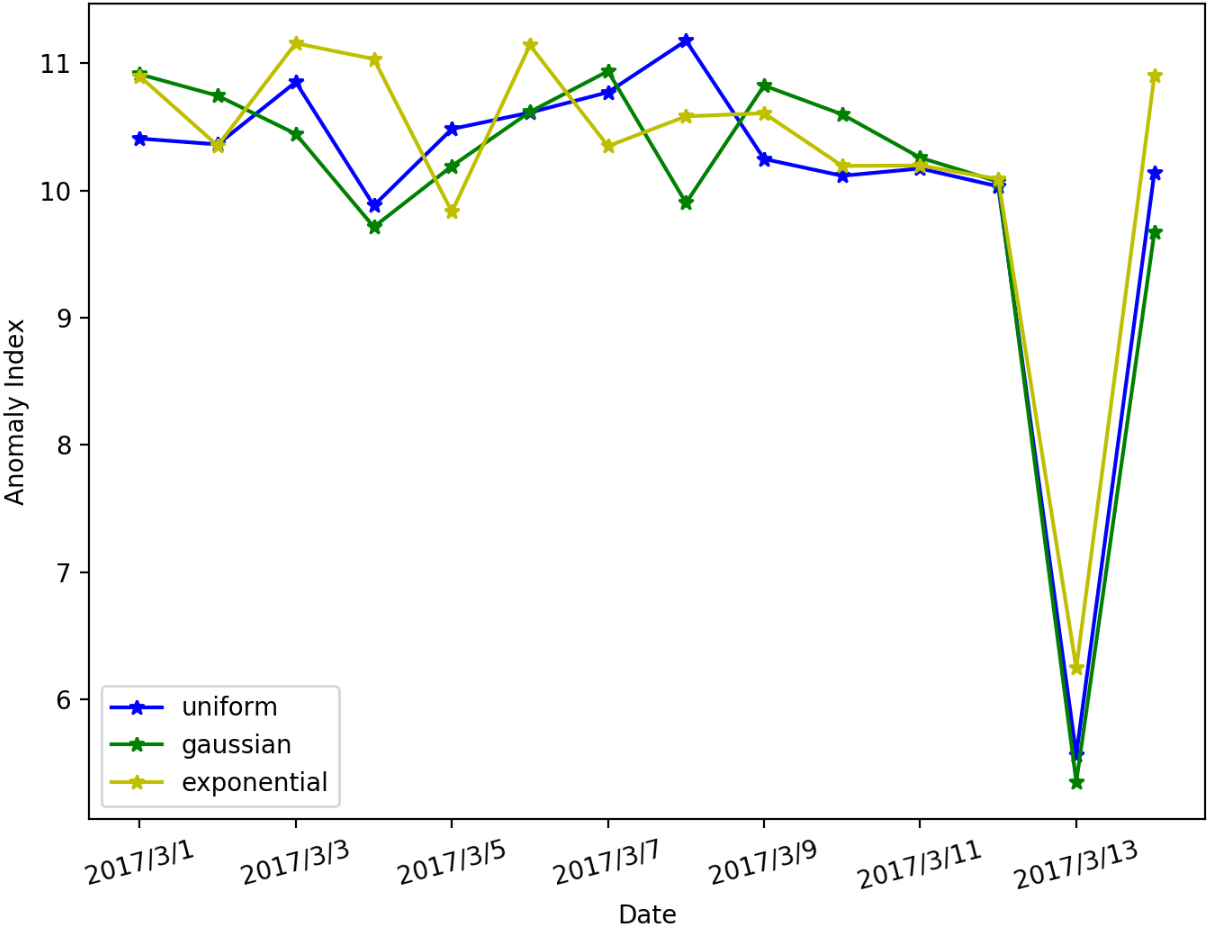}
}
\caption{The $\mathcal{N}_\phi-t$ curve in voltage violation detection.}
\label{fig:Case_voltage_violation_indicator}
\end{figure}

\begin{figure}[!t]
\centering
\begin{minipage}{4.1cm}
\centerline{
\includegraphics[width=.18\textheight,height=.85\textwidth]{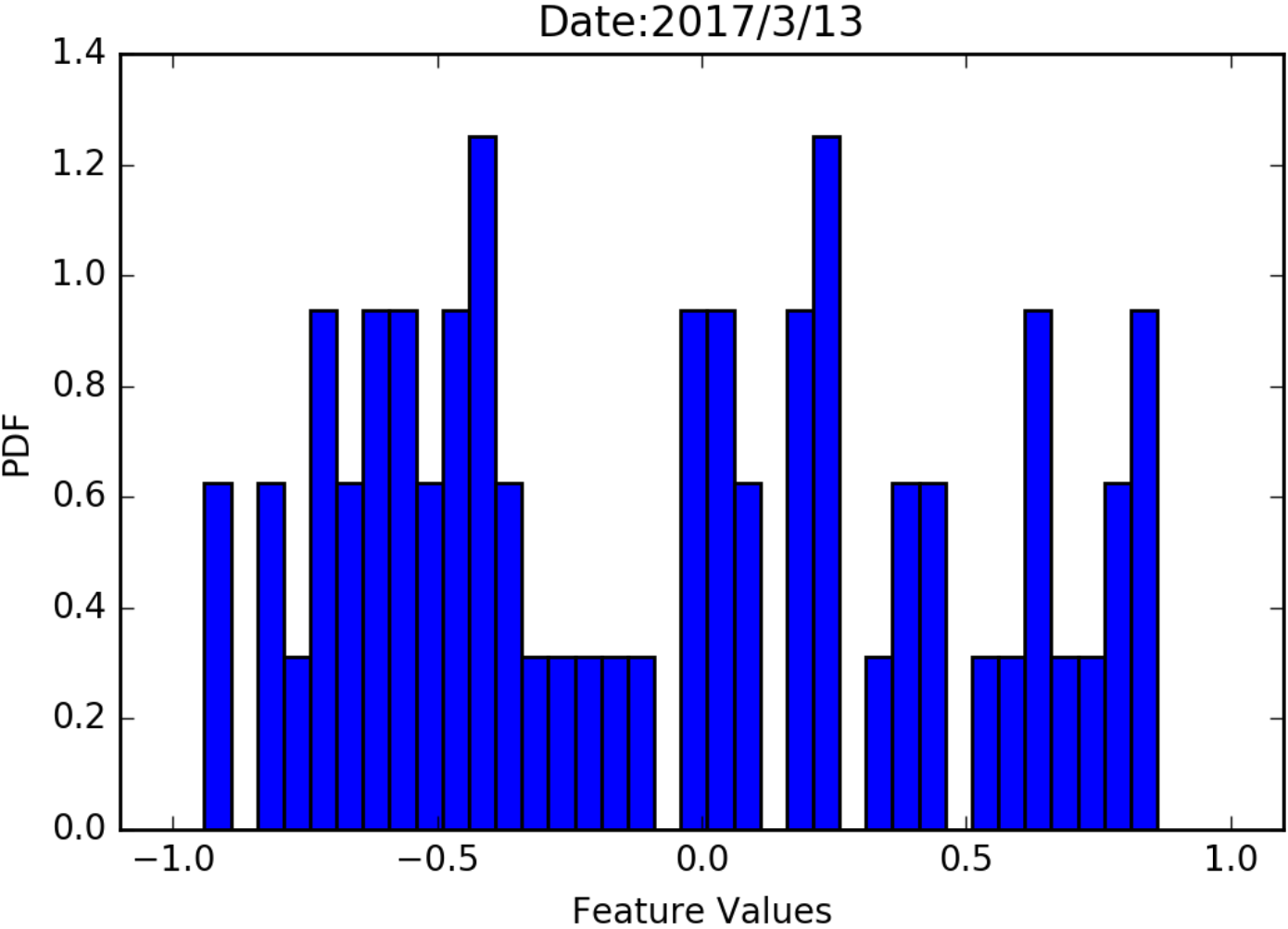}
}
\parbox{5cm}{\small \hspace{1.2cm}(a) March 13th}
\end{minipage}
\hspace{0.2cm}
\begin{minipage}{4.1cm}
\centerline{
\includegraphics[width=.18\textheight,height=.85\textwidth]{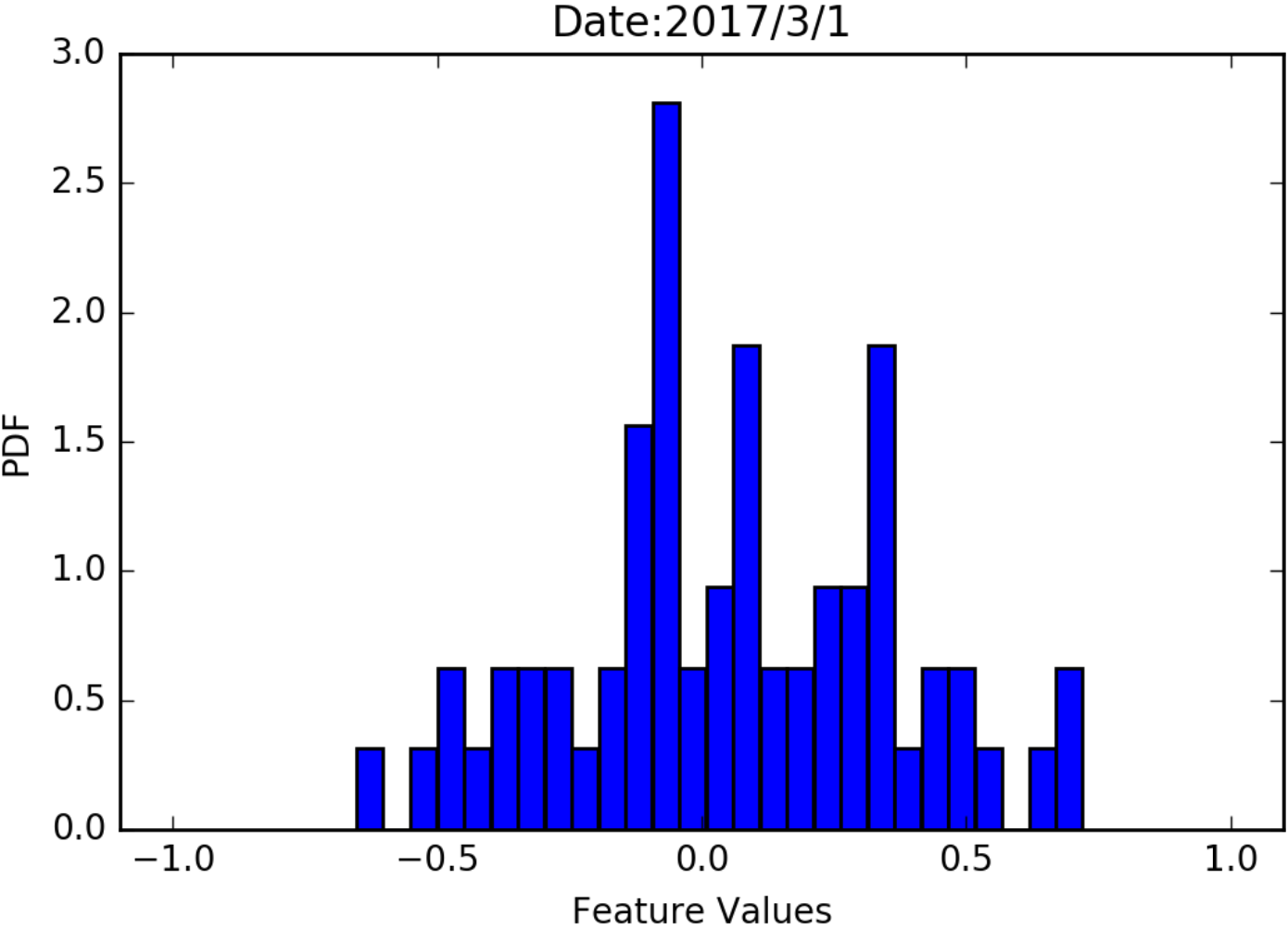}
}
\parbox{5cm}{\small \hspace{1.2cm}(b) March 1st}
\end{minipage}
\caption{The PDF of the extracted features corresponding to abnormal and normal operating states of the feeder line.}
\label{fig:Case_voltage_violation_pdf}
\end{figure}

Figure \ref{fig:Case_voltage_violation_indicator} shows the anomaly detection results when $z$ is sampled from uniform distribution, gaussian distribution, exponential distribution, respectively. From the $\mathcal{N}_\phi-t$ curves, we can obtain:

\uppercase\expandafter{\romannumeral1}. The value of $\mathcal{N}_\phi$ on March 13th is significantly smaller than those on other days, which indicates anomaly occurred on March 13th. The PDFs of the extracted features corresponding to March 13th and other days (such as March 1st) are shown in Figure \ref{fig:Case_voltage_violation_pdf}. It can be observed that the PDFs of the extracted features are different when the feeder operates in different states, i.e., the PDF of the extracted features in normal operating state is more centered.

\uppercase\expandafter{\romannumeral2}. The p values of $\hat{\mathcal{N}_\phi}$ corresponding to the anomaly time for different $z-$sampling distribution were calculated, results of which were $0.1850\%$, $0.2005\%$, $0.2195\%$, respectively. It validates the performance of the proposed approach is almost not affected by the assumption $z-$sampling distribution.

\uppercase\expandafter{\romannumeral3}. The calculated p values of $\hat{\mathcal{N}_\phi}$ on March 13th are smaller than $\frac{\alpha}{2}=\frac{1-97.5\%}{2}=1.25\%$, which indicates the feeder operates in emergency state and it needs to be further analyzed.

\begin{figure}[!t]
\centerline{
\includegraphics[width=2.5in]{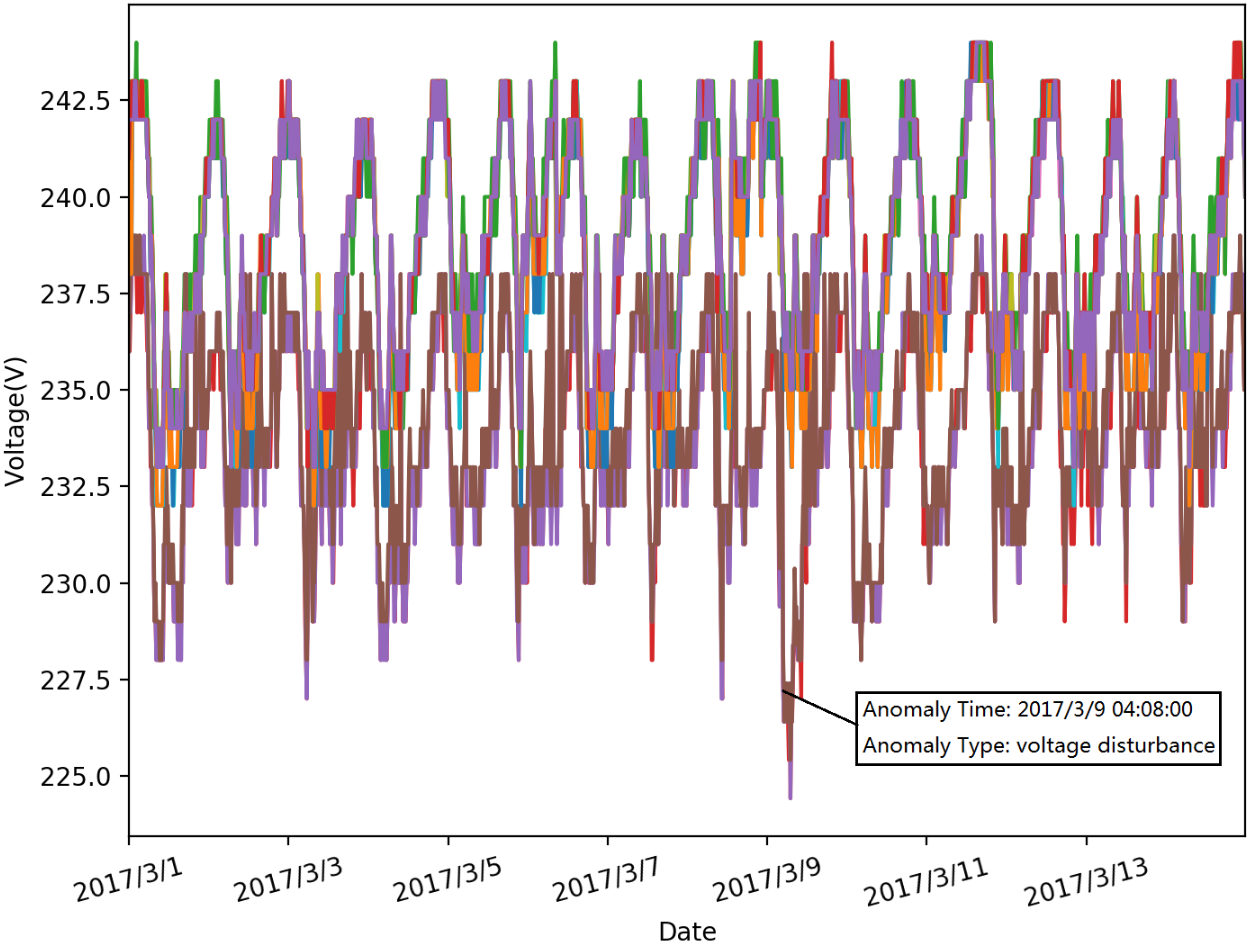}
}
\caption{The real-world online monitoring data with anomaly time and type labelled. The anomaly time is 2017/3/9 04:08:00 and the anomaly type is voltage disturbance.}
\label{fig:case_voltage_disturbance_data}
\end{figure}

2) Case Study on Voltage Disturbance: Voltage disturbance is an complex anomaly type in distribution networks, which is random in magnitude and could involve sporadic bursts as well. It may be caused by short circuit fault, sudden load change, or connection of distribution generation, etc. In this case, the performance of the proposed approach is tested by assessing the operational risk of one feeder line suffering from voltage disturbance. The feeder contained $7$ distribution transformers and the online monitoring data were sampled during 2017/3/1 00:00:00 $\sim$ 2017/3/14 23:45:00, thus a $21\times 1344$ data matrix was formulated. The data with anomaly time and type labelled are shown in Figure \ref{fig:case_voltage_disturbance_data}. The moving window's size was $21\times 96$, the number of neurons in each hidden layer of $D$/$E$ were $1120,672,256$, and the number of neurons in each hidden layer of $G$ were $256,672,1120$. The other parameters were set the same as in the above case.
\begin{figure}[!t]
\centerline{
\includegraphics[width=2.5in]{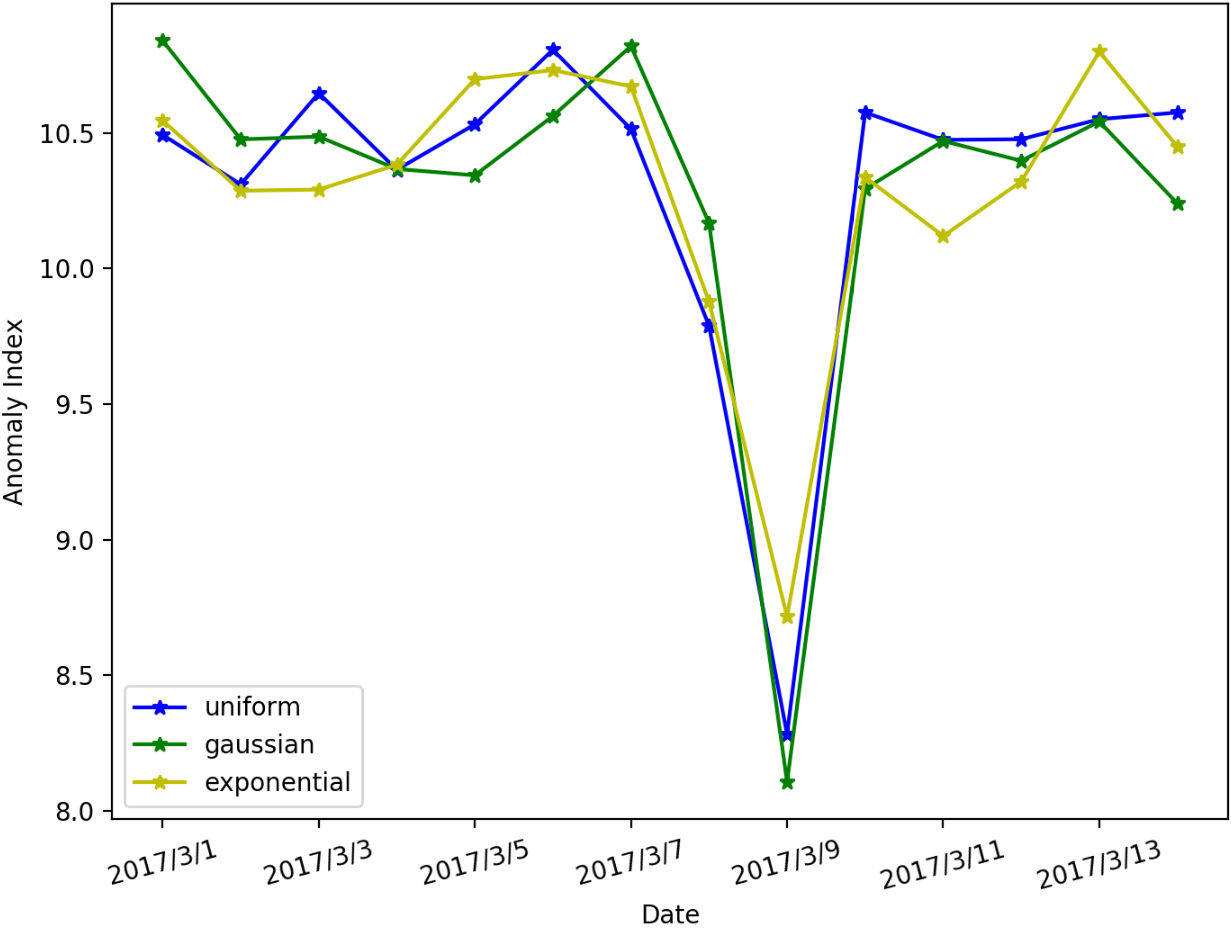}
}
\caption{The $\mathcal{N}_\phi-t$ curve in voltage disturbance detection.}
\label{fig:case_voltage_disturbance_indicator}
\end{figure}

\begin{figure}[!t]
\centering
\begin{minipage}{4.1cm}
\centerline{
\includegraphics[width=.18\textheight,height=.85\textwidth]{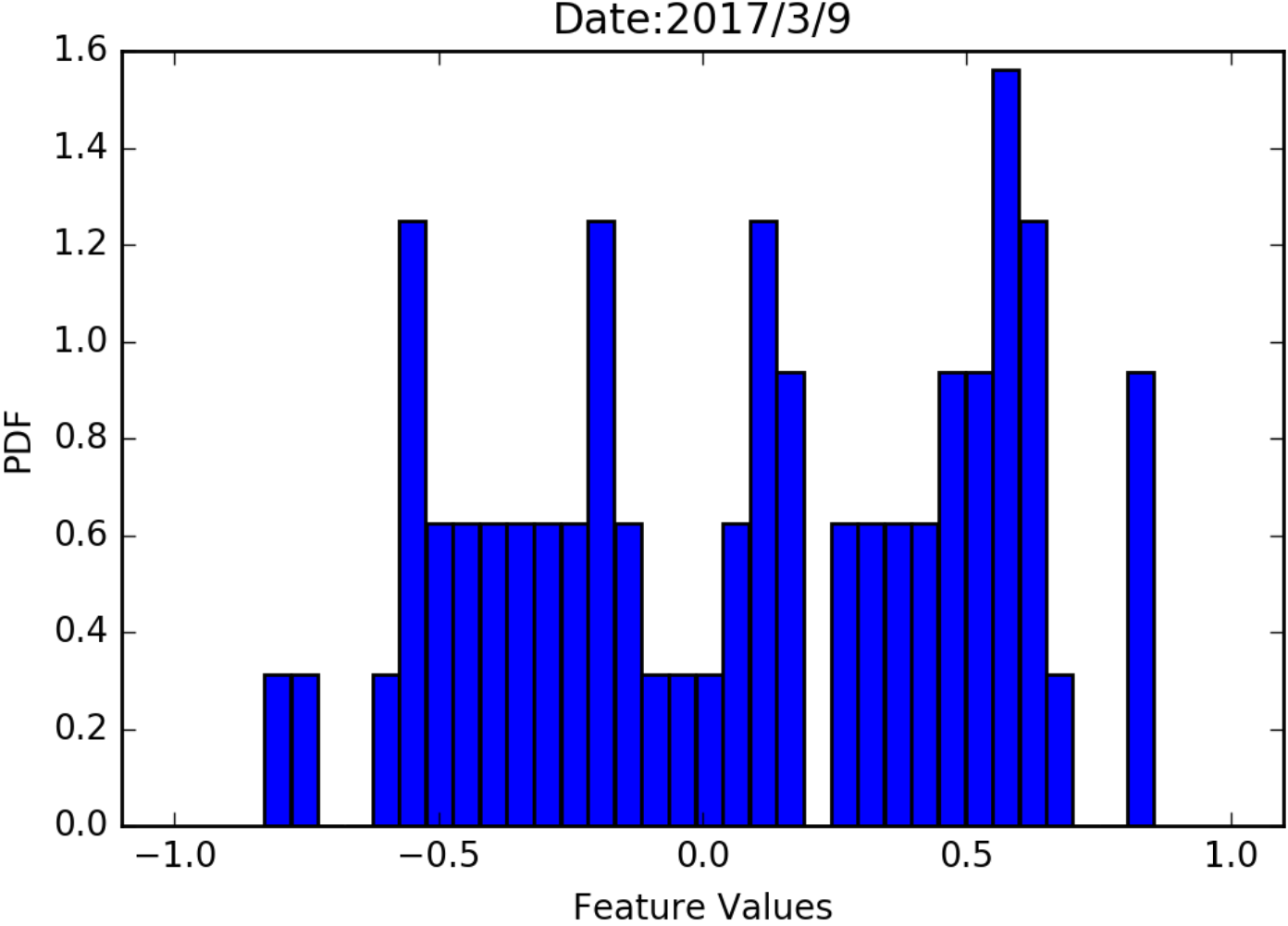}
}
\parbox{5cm}{\small \hspace{1.2cm}(a) March 9th}
\end{minipage}
\hspace{0.2cm}
\begin{minipage}{4.1cm}
\centerline{
\includegraphics[width=.18\textheight,height=.85\textwidth]{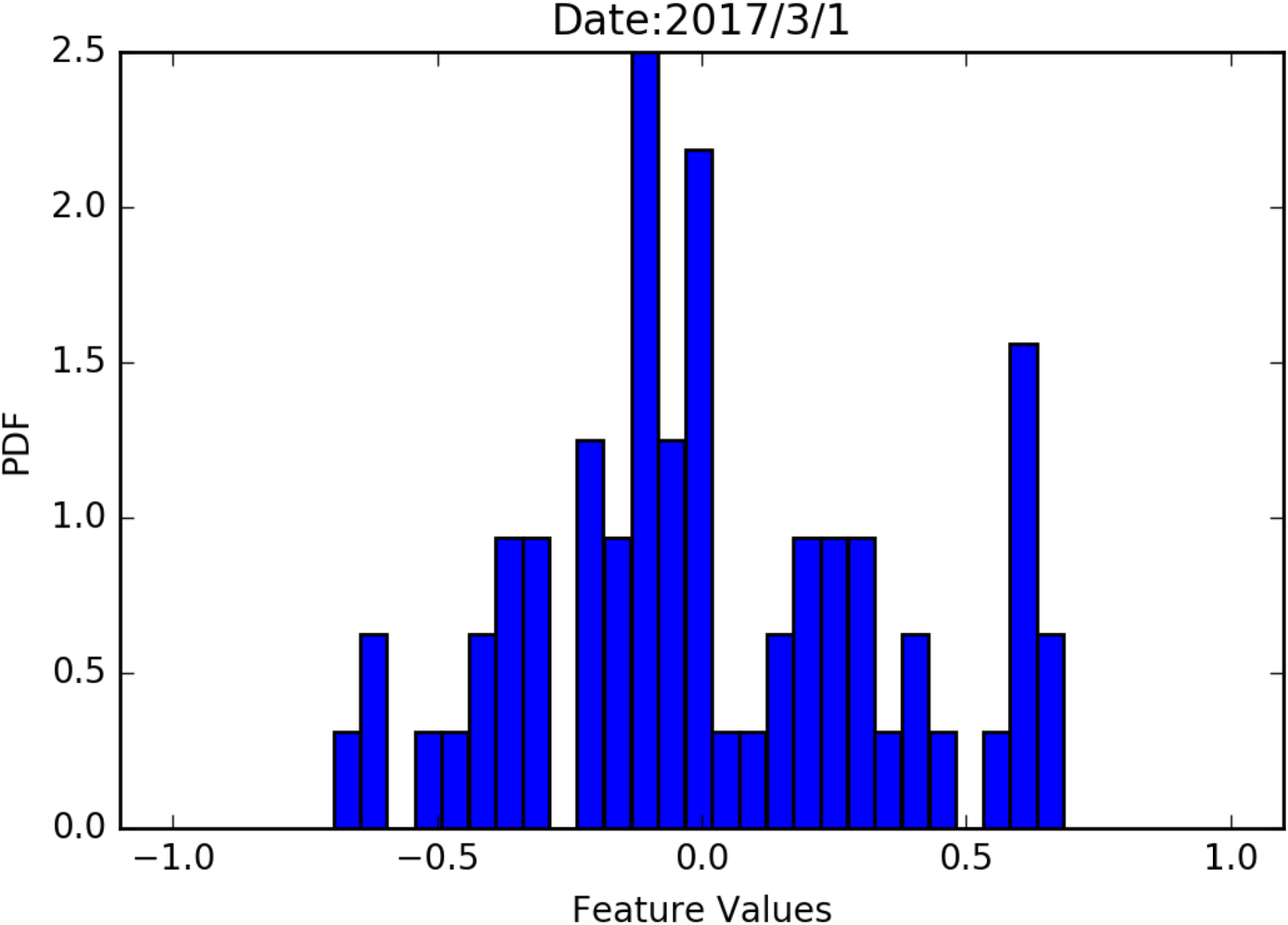}
}
\parbox{5cm}{\small \hspace{1.2cm}(b) March 1st}
\end{minipage}
\caption{The PDF of the extracted features corresponding to abnormal and normal operating states of the feeder line.}
\label{fig:Case_voltage_disturbance_pdf}
\end{figure}

Figure \ref{fig:case_voltage_disturbance_indicator} shows the anomaly detection results corresponding to different $z-$sampling distributions. From the $\mathcal{N}_\phi-t$ curves, we can obtain:

\uppercase\expandafter{\romannumeral1}. The value of $\mathcal{N}_\phi$ on March 9th is smaller than those on other days, which indicates the latent anomaly is accurately detected. The PDFs of the extracted features corresponding to abnormal and normal feeder operating states are shown in Figure \ref{fig:Case_voltage_disturbance_pdf}. It can be observed that, the PDF of the extracted features in normal operating state is more centered.

\uppercase\expandafter{\romannumeral2}. For each $\mathcal{N}_\phi-t$ curve, the p values corresponding to the anomaly time were calculated, results of which were $0.235\%$, $0.196\%$, $0.353\%$, respectively. It also validates the performance of the proposed approach is almost not affected by the assumption of $z-$sampling distribution.

\uppercase\expandafter{\romannumeral3}. The calculated p values of $\hat{\mathcal{N}_\phi}$ on March 9th is smaller than $\frac{\alpha}{2}=1.25\%$, which indicates the feeder operates in emergency state and it deserves to be further analyzed.

3) Comparison with Other Existing Approaches: We further compare the proposed approach with other existing approaches in accuracy and efficiency by assessing the operational risks of feeder lines suffering from anomalies. Here, the risk levels in Table \ref{Tab:state_classification} are simplified as abnormal (emergency state) and normal (the other states). Anomaly detection techniques based on deep autoencoders (DAE) \cite{lee2018time}\cite{barbeiro2015exploiting}, principal component analysis (PCA) \cite{xie2014dimensionality}, spectrum analysis (SPA) \cite{Shi2018Incipient}, or threshold analysis (THA) \cite{xiao2017operation} have been well studied. In order to make a full comparison with the other existing techniques, we analyzed $180$ feeder lines with $250$ anomaly records during 2017/3/1 00:00:00 $\sim$ 2017/4/30 23:45:00. Here, voltage violation and fluctuation were considered as anomaly items. For DAE, PCA, and SPA, the moving window's size was $P\times 96$, the moving step size was $96$ and the test function was $\phi (\lambda)=-\lambda$ln$(\lambda)$. For THA, the anomaly index was defined as
\begin{equation}
\label{Eq:tha_indicator}
\begin{aligned}
  P(\text{A})=\frac{\sum\limits_{j=1}^{n(\text{A})}t(\text{A})_j}{T'}
\end{aligned},
\end{equation}
where $T'$ is the total number of sampling times, $t(\text{A})_j$ is the duration for each abnormal state (i.e., the voltage exceeds the upper or lower limit), $n(\text{A})$ is the number of abnormal states, and $P(\text{A})\in [0,1]$. In the experiments, the optimal parameters involved in each detection approach were tested, and they were set as in Table \ref{Tab: Case6_parameter}.
\begin{table}[!t]
\caption{Parameter Settings Involved in the Detection Approaches.}
\label{Tab: Case6_parameter}
\centering
\footnotesize
\begin{tabular}{p{1.8cm}p{5.8cm}}   
\toprule[1.0pt]
\textbf{Approaches} & \textbf{Parameter Settings} \\
\hline
\multirow{11}*{BiGANs} & the model depth: 5; \\
~&the number of neurons in each hidden layer of $D$: $\lfloor\{0.5,0.3,0.1\}\times P\times 96 \rfloor$; \\
~&the number of neurons in each hidden layer of $E$: $\lfloor\{0.5,0.3,0.1\}\times P\times 96 \rfloor$; \\
~&the number of neurons in each hidden layer of $G$: $\lfloor\{0.1,0.3,0.5\}\times P\times 96 \rfloor$; \\
~&the feature size: $\lfloor\{0.03\sim 0.05\}\times P\times 96 \rfloor$; \\
~&the number of steps $m$ applied to $D$: 1; \\
~&the number of iterations $n$ to calculate $V_{\text{avg}}(D,E,G)$: 5; \\
~&the initial learning rate $\eta$: 0.0001; \\
~&the slope of the leak $\beta$ in LReLu: 0.2; \\
~&the dropout coefficient: 0.1; \\
~&the required approximation error $\varepsilon$: 0.0001; \\
\hline
\multirow{8}*{DAE} & the model depth: $4$; \\
~&the number of neurons in each hidden layer of encoder: $\lfloor\{0.6,0.3\}\times P\times 96 \rfloor$; \\
~&the number of neurons in each hidden layer of decoder: $\lfloor\{0.3,0.6\}\times P\times 96 \rfloor$; \\
~&the feature size: $\lfloor0.1\times P\times 96 \rfloor$; \\
~&the initial learning rate: $0.001$;  \\
~&the activation function: $sigmoid$; \\
~&the minimum reconstruction error: $0.00001$;  \\
~&the optimizer: $Adam$.  \\
\hline
\multirow{1}*{PCA} & the contribution rate of top $k$ eigenvalues: $0.95$. \\
\hline
\multirow{1}*{SPA} & the signal-noise-ratio: $500$. \\
\hline
\multirow{3}*{THA} & the lower limit of voltage violation: $0.93$; \\
~&the upper limit of voltage violation: $1.07$; \\
~&the anomaly threshold $P_{th}$: $0.001$. \\
\hline
\end{tabular}
\end{table}

In order to compare the detection performances of different approaches, $true\; detection\; rate\; (TDR)$ and $false\; alarm\; rate\; (FAR)$ are used to measure the performance of each method. The $TDR$ and $FAR$ are defined as
\begin{equation}
\label{Eq:performance_detection}
\begin{aligned}
  &TDR = \frac{N_{cr}}{N_{gt}} \\
  &FAR = \frac{N_{al}-N_{cr}}{N_{al}}
\end{aligned},
\end{equation}
where $N_{cr}$ is the number of anomalies that are correctly detected, $N_{gt}$ denotes the number of ground-truth anomalies, and $N_{al}$ is the number of all detected alarms. The higher the $TDR$ and the smaller the $FAR$, the better detection performance of one approach. Meanwhile, in order to compare the efficiency of different approaches, the $average\; calculation\; time\;(ACT)$ for each $96$ sampling times (i.e., the moving window's width) was counted. The experiments were conducted on a server with $2.60$ GHz central processing unit (CPU) and $8.00$ GB random access memory (RAM). The comparison results are shown in Table \ref{Tab: Case6_comparison}.
\begin{table}[!t]
\caption{Comparison Results of Different Approaches.}
\label{Tab: Case6_comparison}
\centering
\footnotesize
\begin{tabular}{cccc}   
\toprule[1.0pt]
\textbf{Methods} & $\bm{TDR}$($\%$) & $\bm{FAR}$($\%$) & $\bm{ACT}$(s) \\
\hline
BiGANs & 76.80 & 13.90 & 4.235 \\
DAE & 69.60 & 30.95 & 1.856 \\
PCA & 53.60 & 23.86 & 0.587 \\
SPA & 56.40 & 35.02 & 0.790 \\
THA & 42.80 & 29.17 & 0.314 \\
\hline
\end{tabular}
\end{table}

It can be observed that the proposed BiGANs-based approach has the highest $TDR$ and the smallest $FAR$, which indicates it outperforms DAE, PCA, SPA and THA in anomaly detection performance. The reasons are:
\begin{itemize}
\item THA uses the simple statistical features that often are not well generalized, which makes it impossible to detect the latent anomalies.
\item SPA makes an assumption that the input data follows a certain distribution and the entries of the data matrix are independently and identically distributed. Besides, SPA is an anomaly detection approach based on correlation analysis, which is not sensitive to the amplitude variation of the data.
\item PCA is a linear dimension reduction approach and the optimal parameter measuring the contribution rate of top $k$ eigenvalues is hard to find for all data segments.
\item DAE is a nonlinear generalization of PCA and it is vulnerable to the random fluctuations for the reason of simple network structure and learning algorithm.
\end{itemize}

The proposed approach overcomes the shortcomings of the other existing approaches by using complex $D,E,G$ to learn the features of the input data in an adversarial way. Meanwhile, it is noted that the proposed approach has the highest $ACT$ (i.e., $4.235$s) for the reason of complex network structure and learning algorithm, which indicates the worst detection efficiency compared with the other approaches. However, considering the online monitoring data in the researched network is sampled every 15 minutes, the proposed approach is practical for the online operational risk analysis. Moreover, with the development of graphics processing unit (GPU) and field-programmable gate array (FPGA) techniques, the computational efficiency will be improved greatly.

\section{Conclusion}
\label{section: conclusion}
This paper proposes a new unsupervised approach to realize the operational risk assessment in distribution networks. The proposed approach is capable of mining the hidden structure of the real data and automatically learning the most representative features of the data in an adversarial way. By analyzing the distribution of the extracted features, a statistical index is calculated to indicate the data behavior. The standard form of the index is experimentally proved to approximate the t distribution. Furthermore, the operational risks of feeders are divided into emergency, high risk, preventive and normal by the defined intervals of the confidence level for the population mean of the standardized index, which makes it possible for the quantitative risk assessment.

Cases on the synthetic data offer guidelines for the parameter selections of the proposed approach, including simple $z-$sampling distribution, moderate model depth and feature size. Cases on the real-world online monitoring data indicate the proposed approach can improve the risk assessment accuracy a lot compared with the other existing techniques. However, the computational burden is increased for the complex network structure and learning algorithm of the approach. In view of the outstanding advantages in assessment performance, the proposed approach can serve as a primitive for analyzing the spatio-temporal data in distribution networks.





\small{}
\bibliographystyle{IEEEtran}
\bibliography{helx}

\end{document}